\documentclass[10pt,twocolumn,letterpaper]{article}

\usepackage[pagenumbers]{iccv} % To force page numbers, e.g. for an arXiv version

\definecolor{iccvblue}{rgb}{0.21,0.49,0.74}
\usepackage[pagebackref,breaklinks,colorlinks,allcolors=iccvblue]{hyperref}

\usepackage{graphicx}
\usepackage{amsmath}
\usepackage{amssymb}
\usepackage{mathtools}
\usepackage{amsthm}
\usepackage[ruled,vlined]{algorithm2e}
\usepackage{algorithmic}
\usepackage{multirow}
\usepackage{comment}
\usepackage{color, colortbl, soul}
\definecolor{Gray}{gray}{0.9}
\usepackage{booktabs}
\usepackage{tabularx, booktabs, makecell, caption}
\usepackage{siunitx}
\usepackage{multicol}
\usepackage{subcaption}
\usepackage{lipsum,calc}
\usepackage{xspace}
\usepackage{pifont}

\theoremstyle{plain}
\newtheorem{theorem}{Theorem}

\newtheorem{lemma}{Lemma}

\theoremstyle{definition}
\theoremstyle{remark}

\SetAlgoNoEnd                 % for, if 등에 end 텍스트 출력 X
\title{FedWSQ: Efficient Federated Learning with Weight Standardization and Distribution-Aware Non-Uniform Quantization}

\author{
Seung-Wook Kim\footnotemark[1]~~$^1$, Seongyeol Kim\footnotemark[1]~~$^1$, Jiah Kim\footnotemark[1]\thanks{indicates equal contribution.}~~$^1$, Seowon Ji\footnotemark[2]~~$^2$, and Se-Ho Lee\footnotemark[2]\thanks{Corresponding authors.}~~$^3$\\
$^1$Pukyong National University \quad $^{2}$Konkuk University \quad $^{3}$Jeonbuk National University
}

\begin{document}
\maketitle

% !TEX root = ../main.tex

%%%%%%%%% ABSTRACT
\begin{abstract}

Federated learning (FL) often suffers from performance degradation due to key challenges such as data heterogeneity and communication constraints.
To address these limitations, we present a novel FL framework called FedWSQ, which integrates weight standardization~(WS) and the proposed distribution-aware non-uniform quantization~(DANUQ).
WS enhances FL performance by filtering out biased components in local updates during training, thereby improving the robustness of the model against data heterogeneity and unstable client participation. In addition, DANUQ minimizes quantization errors by leveraging the statistical properties of local model updates. As a result, FedWSQ significantly reduces communication overhead while maintaining superior model accuracy.
Extensive experiments on FL benchmark datasets demonstrate that FedWSQ consistently outperforms existing FL methods across various challenging FL settings, including extreme data heterogeneity and ultra-low-bit communication scenarios. 
The source code is available at our project page\footnote{\url{https://github.com/Seongyeol-kim/FedWSQ}}.
\vspace{-7mm}
\end{abstract}

% !TEX root = ../main.tex

\section{Introduction}
\label{sec:introduction}
In large-scale machine learning, centralized approaches often raise privacy concerns because sensitive data from edge devices must be collected on a central server for model training. 
Federated learning~(FL)~\cite{mcmahan2017communication} enables distributed devices to collaboratively train a shared global model without sharing raw data.
A fundamental FL method, FedAvg, was proposed by McMahan~\etal~\cite{mcmahan2017communication}. 
In this approach, each client trains its model using the local data, and then the server collects and averages these local updates to form a unified global model.

Several studies on FL have verified its practical effectiveness~\cite{stich2018local,yu2019parallel,wang2021cooperative,stich2019error,basu2020qsparse,yang2021achieving,li2019convergence}.
However, real-world FL methods often face three key challenges~\cite{khaled2019first,karimireddy2020scaffold,le2024exploring,luping2019cmfl,wen2023survey}: 1) \textit{data heterogeneity}, where clients possess non-independent and identically distributed~(non-\textit{i.i.d.}) data; 2) \textit{partial client participation}, where only a subset of clients contributes to a global model update during each training round due to communication constraints; 3) \textit{communication bottlenecks}, where limited communication bandwidth and high transmission costs hinder efficient model aggregation.
These challenges worsen local gradient divergence, slow down the global model convergence, and ultimately degrade FL performance in real-world applications.

To address these limitations, we propose FedWSQ, an advanced FL framework that combines weight standardization~(WS) with our proposed distribution-aware non-uniform quantization~(DANUQ) to improve both learning stability and communication efficiency.
WS is a plug-and-play technique that standardizes the weight vectors of convolutional or linear layers to stabilize the learning process of a neural network~\cite{qiao2019ws}.
In FedWSQ, WS plays a crucial role in mitigating client drift by filtering out gradient components that contribute to overfitting during local training. This leads to improved generalization across heterogeneous clients and significantly enhances FL performance.
Although WS accelerates global convergence, the communication cost per round remains a major bottleneck.
To address this, FedWSQ integrates WS with DANUQ, a novel quantization strategy that leverages a standard normal distribution prior to minimize quantization errors, reducing communication overhead while preserving model performance.
Even under ultra-low-bit quantization, the proposed FedWSQ achieves superior performance, consistently outperforming state-of-the-art~(SOTA) FL methods across various datasets and FL settings. 

Our main contributions are summarized as follows.
\begin{itemize}
	
    \item[$\bullet$] We propose an effective application of WS in FL, enhancing model convergence and stability under data heterogeneity and limited client participation. By implicitly performing a gradient filtering process, WS mitigates client drift while also providing a regularization effect that stabilizes training and improves generalization.

    \item[$\bullet$] We introduce DANUQ, a novel quantization method that employs a fixed quantization function, leveraging the statistical properties of local model parameter updates~(LMPUs) to minimize quantization errors under parameter-distribution priors. 
    By efficiently compressing LMPUs while preserving essential information, the proposed DANUQ significantly reduces communication overhead without noticeable performance degradation.

    \item[$\bullet$] Extensive experiments show that FedWSQ significantly enhances performance across diverse settings. 
    Even with an average of 2.3 bits per parameter, FedWSQ improves performance by over 5\% on Tiny-ImageNet under a highly heterogeneous setting compared to SOTA FL methods.

\end{itemize}

% !TEX root = ../main.tex

\begin{figure*}[t]
\centering
\includegraphics[width=0.8\linewidth]{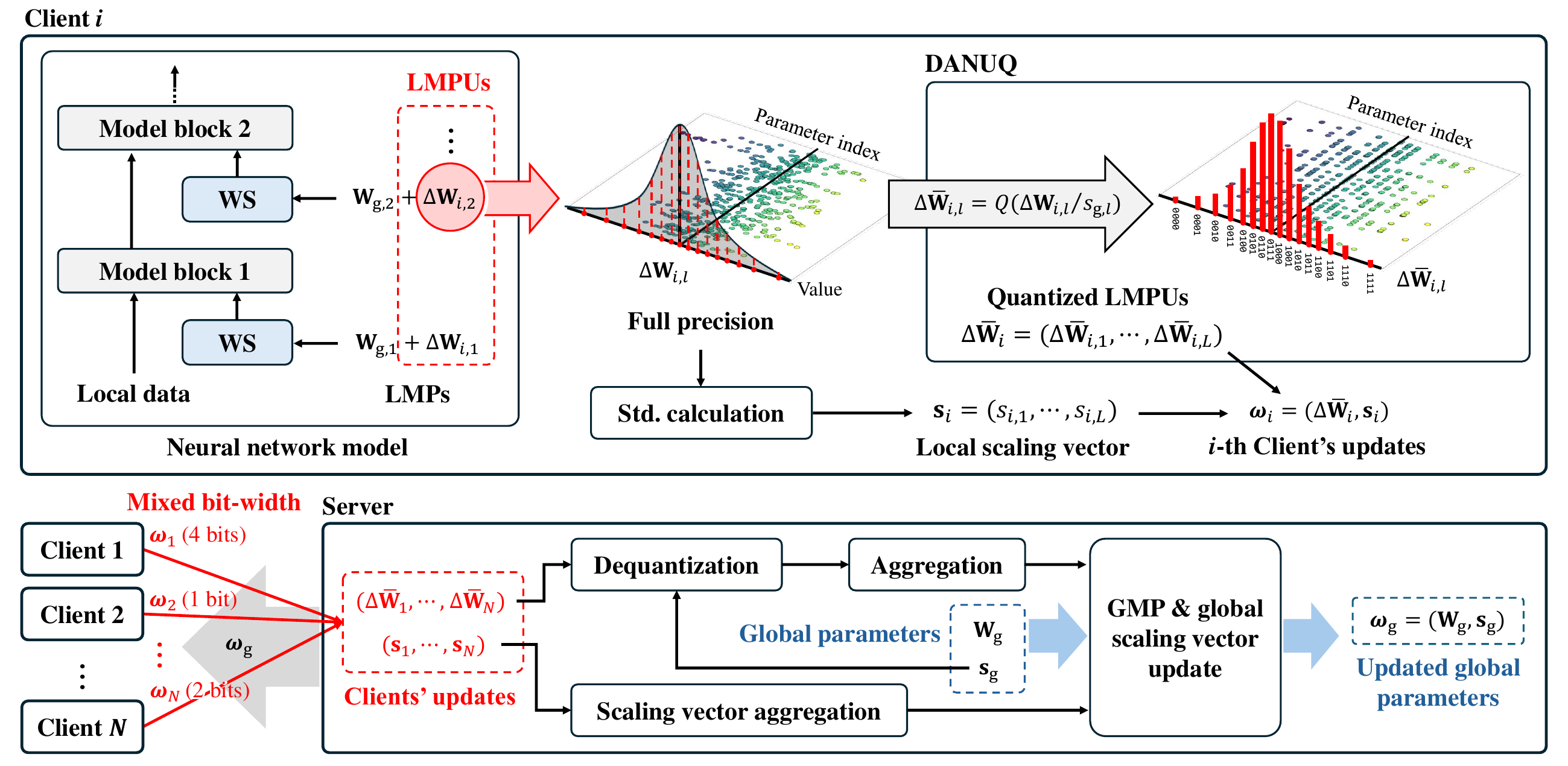}
\vspace{-2mm}
\caption{The overall framework of FedWSQ, which integrates WS and DANUQ for efficient FL. Clients transmit LMPUs quantized by the DANUQ along with scaling vectors to the server, where dequantization and aggregation are performed to update the global parameters.}
\label{fig:framework}
\vspace{-3mm}
\end{figure*}

\section{Related Work and Preliminaries} 
\label{sec:related}

\paragraph{Federated learning~(FL)}
FL is a decentralized approach that enables collaborative training of a global model.
Using distributed devices, FL ensures efficient learning while preserving data privacy. 
Let $\mathcal{F}_n(\mathbf{w}_\mathrm{g}) \coloneq \mathbb{E}_{\zeta_n \sim \mathcal{D}_n} \left[ \mathcal{F}_n(\mathbf{w}_\mathrm{g};\zeta_n) \right]$ be the local loss function of the client $n \in \left[ N \right]$ with a local data distribution $\mathcal{D}_n$, where $\mathbf{w}_\mathrm{g}$ is the global model parameter~(GMP) vector. The training objective function of FL~\cite{mcmahan2017communication} is formulated as follows: 
\vspace{-1mm}
\begin{equation}
\label{global_objective}
    \underset{\mathbf{w}_\textrm{g}}{\operatorname{min}}\Big\{\mathcal{F}(\mathbf{w}_\mathrm{g}) := \sum_{i \in \mathcal{S}} h_i \mathcal{F}_i(\mathbf{w}_\mathrm{g})\Big\},
\vspace{-2mm}
\end{equation}
where $h_i$ is the weight assigned to the $i$-th device, such that $h_i \geq 0$, $\sum_i h_i = 1$, and $h_i \propto \left\vert \mathcal{D}_i \right\vert$, and $\mathcal{S} \subseteq [N]$ is a subset of participating clients in the current communication round. At each round $t \in \left[ T \right]$, the server broadcasts the GMPs $\mathbf{w}_\textrm{g}^{t-1}$ to the participating client $i$. In FedAvg~\cite{mcmahan2017communication}, the $i$-th client performs $K$ steps of local training to obtain the local model parameter~(LMP) vector,  $\mathbf{w}_{i}^{t}$, and transmits its LMP updates~(LMPUs), $\Delta \mathbf{w}_i^t \coloneq \mathbf{w}_{i}^{t} - \mathbf{w}_\textrm{g}^{t-1}$, to the server. Finally, the server aggregates these LMPUs through weighted averaging to derive the updated GMPs.

In real-world FL scenarios, unstable learning dynamics and slow convergence often occur due to non-\textit{i.i.d.} data distributions and limited communication bandwidth. To address these issues, FedProx~\cite{li2020federated} penalizes LMPUs by adding a proximal term in the local objective to reduce the gap between the global and local loss functions. FedAvgM~\cite{hsu2019measuring} integrates momentum into the global update process to mitigate variance in model aggregation. FedAdam~\cite{reddi2021adaptive} introduces an adaptive optimizer that improves global and local model convergence, thereby accelerating overall training progress. FedDyn~\cite{acar2021federated} presents dynamic regularization for each device to diminish deviations in the LMPUs caused by data heterogeneity. FedSmoo~\cite{sun2023dynamic} applies constraints that prevent sharp fluctuations in local updates to generalize FL performance by ensuring global consistency.
Recently, FedACG~\cite{kim2022multi} was proposed to improve convergence by broadcasting a global model with a lookahead gradient and aligning local models with the shared global model. 

\paragraph{Network quantization}
As neural network architectures have grown larger in scale, quantization has become essential for the memory optimization and acceleration of operations~\cite{gray1998quantization, lin2016fixed,Nagel2021quantization, Jacob2018quantization,LSQ,jung2019qil,zhu2017ttq}. Let us consider a model using $B$-bit representation. A typical quantization function $\mathcal{Q}$ mapping a full-precision value $x$ to a quantization level~(QL) $q$ can be written as
\begin{equation} 
q = \mathcal{Q}(x/s),
\label{eq:quant_func}
\end{equation}
where $s$ is a scaling factor, and each QL is matched with a $B$-bit integer from $\{0,\dots,2^B-1\}$.
Note that $\mathcal{Q}(\cdot)$ assigns an input value to one of $2^B$ QLs through nearest-neighbor clustering. The primary goal of effective quantization is to determine the proper scaling factor and a set of QLs to minimize the quantization errors. 

Quantization methods can be roughly categorized into two approaches: uniform quantization~(UQ) and non-uniform quantization~(NUQ).
UQ divides the input range into $2^B$ equal intervals and assigns a representative value as a QL to each interval~\cite{gray1998quantization, lin2016fixed}.
In contrast, NUQ adjusts the interval widths based on the prior data distribution ~\cite{PACT, LQ-Nets, QLoRA}. 
By allocating finer intervals in dense data regions and broader intervals in sparse regions, NUQ preserves critical information more effectively than UQ.
For the scaling factor $s$, both approaches commonly employ the \textit{absmax} method~\cite{Nagel2021quantization, Jacob2018quantization} to determine the dynamic range for quantization.
Furthermore, to enhance data adaptability, researchers have explored methods for dynamically learning quantization parameters (\textit{e.g.}, scaling factors and QLs)~\cite{LSQ,jung2019qil,zhu2017ttq}.
During backpropagation, these parameters are updated to reduce quantization error and better align with the underlying data distribution. 

\vspace{-3mm}
\paragraph{Quantized FL}
In FL, communication bottlenecks arise primarily during uplink transmission, as each participating client must send its individual LMPUs to the central server. Unlike downlink transmission, where a single GMP is broadcast to multiple clients, uplink communication involves substantial data transfer. This problem is exacerbated in resource-constrained communication environments, such as mobile and IoT networks, where upload bandwidth is severely limited~\cite{reisizadeh2020fedpaq,Chen2021fedHQ+}. To overcome this challenge, several studies have focused on compressing model updates.
In~\cite{FLGOOGLE}, quantization and sparsification are combined to reduce the number of transmitted parameters.
FedPAQ~\cite{reisizadeh2020fedpaq} employs a probabilistic rounding algorithm to prevent an excessive concentration of identical QLs. Meanwhile, FedHQ+~\cite{Chen2021fedHQ+} assigns different weight factors to clients for global updates based on their individual quantization errors. 
Although these methods significantly improve communication efficiency, they often suffer from performance degradation under extremely low bit-width conditions. In this paper, we introduce FedWSQ, a novel quantization framework that is designed to maintain robust model performance even at ultra-low bit-width.

% !TEX root = ../main.tex
\vspace{-2mm}
\section{Proposed Method}
\label{sec:proposed_method}

Figure~\ref{fig:framework} illustrates the overall framework of the proposed FedWSQ, which employs WS and DANUQ for efficient FL. Each client has a standard neural network model consisting of multiple model blocks, where WS is applied to mitigate the impact of data heterogeneity between clients. 
After local training, the full-precision values of LMPUs are converted to a $B$-bit representation using the global scaling vectors and carefully designed QLs. These quantized LMPUs and their corresponding scaling vectors are transmitted to the server. The server then dequantizes the received data to higher-precision values and aggregates them to update the GMPs. In addition, the global scaling vectors are refined to ensure consistent scaling across different clients for DANUQ. Using the proposed DANUQ, FedWSQ enhances communication efficiency without sacrificing model performance in challenging FL scenarios.

\subsection{Weight standardization~(WS)}
\label{sec:ws}

WS~\cite{qiao2019ws} standardizes each weight vector of a convolution or linear layer to enhance the learning process. By ensuring consistent parameter distributions during training, WS contributes to a more stable gradient flow.
Beyond this general benefit, we argue that the use of WS in FL can address the challenges posed by the distributed nature of FL and data heterogeneity.
Formally, consider the $n$-th client that has a network architecture consisting of $L$ linear or convolutional layers.
Let the $l$-th layer (without bias) map an input $\mathbf{x}_l \in \mathbb{R}^{I_l}$ to an output $\mathbf{y}_l \in \mathbb{R}^{O_l}$, defined by
\vspace{-1mm}
\begin{align}
\mathbf{y}_l &= \mathbf{W}_{n,l}^T\mathbf{x}_l \\
&\coloneq \begin{bmatrix}
\mathbf{w}_{n, m_{1}^l} & \mathbf{w}_{n, m_{2}^l} & \cdots & \mathbf{w}_{n, m_{O_l}^l}
\end{bmatrix}^T \mathbf{x}_l, %\\
%&\eqcolon %\mathbf{W}_l\mathbf{x}_l,
\label{eq:conv}
\end{align}
where $\mathbf{w}_{n,m}$, for $m \in \mathcal{M}_l \coloneq \{m_1^l,\dots,m_{O_l}^l\}$, is the weight vector.\footnote{Strictly speaking, convolution involves spatial structures aggregating information from local receptive fields. For simplicity, we omit spatial considerations and treat the convolution as a linear transformation. This abstraction can be easily generalized to multidimensional tensor structures.} 
In WS, each $\mathbf{w}_{n,m}$ is standardized before being used within the layer.
% We denote the pre-standardized parameter~(PSP) vector, \textit{i.e.}, the input of WS, as $\mathbf{w}_{n,m}$, and the weight-standardized parameter~(WSP) vector, \textit{i.e.}, the output of WS, as $\tilde{\mathbf{w}}_{n,m}$, respectively. 
We denote the pre-standardized parameter~(PSP) vector as $\mathbf{w}_{n,m}$, and the weight-standardized parameter~(WSP) vector as $\tilde{\mathbf{w}}_{n,m}$, respectively. 
Then, the WSP vector $\tilde{\mathbf{w}}_{n,m}$ is obtained by
\vspace{-1mm}
\begin{align}
\tilde{\mathbf{w}}_{n,m} &= \frac{\rho}{\sigma(\mathbf{w}_{n,m})} \left(\mathbf{I}-\mathbf{P}_\mathbf{1} \right){\mathbf{w}_{n,m}},
\label{eq:ws}
\vspace{-1mm}
\end{align}
where $\mathbf{1} \in \mathbb{R}^{I_l}$ is a vector of all ones, $\sigma(\mathbf{w}_{n,m})$ is the standard deviation of the elements in $\mathbf{w}_{n,m}$,
% \footnote{In practice, to avoid division by zero, the standard deviation is stabilized by adding a small constant $\epsilon$, \ie, $\sigma \rightarrow \sqrt{\sigma^2 + \epsilon}$. For simplicity, we omit this adjustment in our discussion.} 
and $\mathbf{P}_{\mathbf{v}}$ represents the projection matrix onto the subspace spanned by the vector $\mathbf{v}$, which is denoted by $\operatorname{span}\{\mathbf{\mathbf{v}}\}$. Here, we introduce a hyper-parameter $\rho$ to control the normalization scale.
Note that we reformulate the WSP computation from~\cite{qiao2019ws} into Eq.~(\ref{eq:ws}) to better understand the gradient filtering process of WS during local training.

\vspace{-2mm}
\paragraph{Gradient filtering by WS}
WS plays an important role in reducing the learning diversity of local models, which is one of the most critical issues in FL.
To understand its impact, we compare the gradient of a loss function $\mathcal{L}$ with respect to the WSP vector $\tilde{\mathbf{w}}_{n,m}$ to that of the PSP vector $\mathbf{w}_{n,m}$.
When WS is applied, the gradient is processed by an additional series of projections to remove specific components that may interfere with effective FL.
Specifically, the gradient with respect to the PSP vector $\mathbf{w}_{n,m}$ is obtained by\footnote{Section~\ref{supp:deriv1} of the supplementary document provides its derivation.}
\begin{align}
\frac{\partial \mathcal{L}}{\partial \mathbf{w}_{n,m}} &= \frac{\rho}{\sigma(\mathbf{w}_{n,m})}\left( \mathbf{I}-\mathbf{P}_{\mathbf{1}} \right) \left( \mathbf{I}-\mathbf{P}_{\tilde{\mathbf{w}}_{n,m}} \right)\frac{\partial \mathcal{L}}{\partial \tilde{\mathbf{w}}_{n,m}}.
\label{eq:backward_proj}
\end{align}
The upstream gradient $\frac{\partial \mathcal{L}}{\partial \tilde{\mathbf{w}}_{n,m}}$ is projected onto two subspaces, $\operatorname{span}\lbrace \tilde{\mathbf{w}}_{n,m} \rbrace^{\perp}$ and $\operatorname{span}\lbrace \mathbf{1} \rbrace^{\perp}$, where $\operatorname{span}\lbrace \mathbf{v} \rbrace^{\perp}$ represents the orthogonal complement of $\operatorname{span}\lbrace \mathbf{v} \rbrace$. 
Since $\tilde{\mathbf{w}}_{n,m} \perp \mathbf{1}$, the upstream gradient
$\frac{\partial \mathcal{L}}{\partial \tilde{\mathbf{w}}_{n,m}}$ is ultimately projected onto $\operatorname{span}\lbrace \tilde{\mathbf{w}}_{n,m}, \mathbf{1} \rbrace^{\perp}$ after the two projections.

\begin{figure}[t]
\centering
\includegraphics[width=\linewidth]{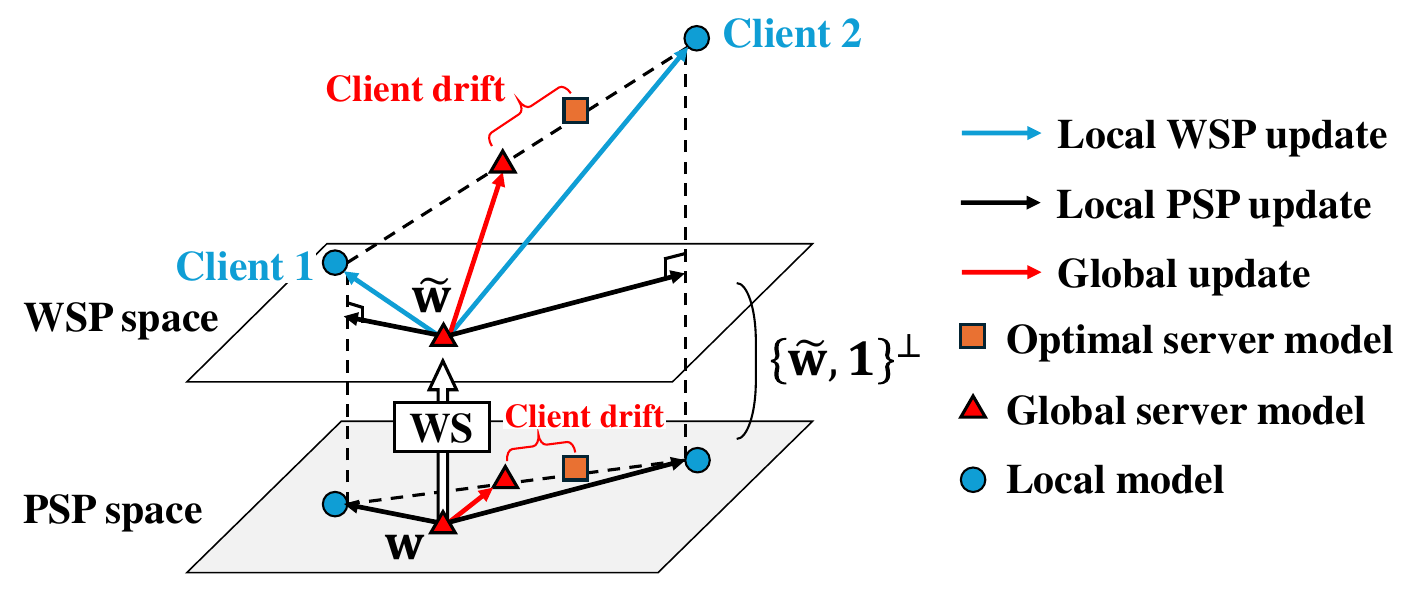}
\vspace{-0.6cm}
\caption{An example of gradient filtering process of WS.}
\label{fig:ws}
\vspace{-0.3cm}
\end{figure}

As illustrated in Figure~\ref{fig:ws}, Eq.~(\ref{eq:backward_proj}) represents a gradient filtering process that applies successive projections to remove undesirable components from the gradient with respect to LMPs.
In FL, data heterogeneity causes local models to overfit their local data, which leads to inconsistent updates across clients.
This inconsistency results in a discrepancy between the GMPs and the optimal parameters, known as client drift.
WS mitigates this issue by projecting the gradient onto $\operatorname{span}\lbrace \tilde{\mathbf{w}}_{n,m}, \mathbf{1} \rbrace^{\perp}$, which effectively reduces the impact of client drift during training.
The first projection eliminates the gradient component aligned with $\tilde{\mathbf{w}}_{n,m}$, which corresponds to the LMPs that are biased due to the non-\textit{i.i.d.} data distribution. The second projection removes the mean component of mini-batch gradients, which can also be biased toward the local data distribution, compounding inconsistencies across clients.
By filtering out both the parameter-aligned and mean components from the local gradient, WS preserves meaningful gradient directions essential for stable global model convergence.
Additionally, this projection-based filtering introduces a regularization effect by constraining the model's deviation from its current state, which can improve the model's generalization ability.

\vspace{-2mm}
\paragraph{Comparison with FedWon}
\label{sec:comp_fedwon}
It is worth noting that FedWon~\cite{zhuang2024fedwon}, a recently proposed FL method, also employs WS to address data heterogeneity in multi-domain FL.
However, a key difference lies in the parameters exchanged between clients and the server, which fundamentally alters the learning dynamics.
FedWon transmits WSP, which explicitly normalizes parameters across clients to improve learning stability.
However, this approach forces all clients' LMPUs to exhibit statistically similar characteristics, which potentially discards client-specific information crucial for local adaptation.
In contrast, FedWSQ transmits PSP, preserving essential local information while implicitly mitigating harmful divergences through the gradient filtering process of WS, as shown in Figure~\ref{fig:ws}.
This design choice enables FedWSQ to achieve a better balance between global stability and local information preservation.
An experimental comparison between FedWon and FedWSQ will be discussed in Section~\ref{ssec:analysis}.

% \begin{figure}[t]
% \centering
% \includegraphics[width=0.8\linewidth]{figures/fitting.pdf}
% \vspace{-0.3cm}
% \caption{Distributions of LMPUs in the 1st and 3rd blocks of ResNet-18, with values normalized by their standard deviation. The red dotted line represents a standard normal distribution.}
% \label{fig:fitting}
% \vspace{-0.3cm}
% \end{figure}

\subsection{Quantization of LMPUs}
\label{sec:quantization}
Neural network quantization often involves learning quantization parameters.
However, in FL, transmitting additional quantization parameters or auxiliary information for each communication round would introduce significant resource overhead. Instead, we propose a fixed quantization function that eliminates the need to transmit additional quantization information, ensuring a lightweight and efficient communication process.
A fundamental assumption in NUQ is that model parameters follow a normal distribution, as demonstrated in prior studies~\cite{zhao2021distribution,dettmers20228bit,wang2025optimizing}. 
Since LMPUs represent the difference between an updated LMP and the previous GMP, they may also exhibit a normal distribution. 
% To empirically validate this assumption, Figure~\ref{fig:fitting} presents histograms of LMPUs in the network model. As indicated by the red dotted line in the figure, the normal distribution well approximates the actual distribution of LMPUs.

\vspace{-2mm}
\paragraph{LMPU scaling}
Most quantization methods adopt the \textit{absmax} strategy to manage the dynamic range of values~\cite{Nagel2021quantization, Jacob2018quantization}.
This approach scales an input tensor to the range $[-1, 1]$ by dividing each element by the maximum absolute value of the tensor. Within this constrained range, conventional quantization methods determine QLs and map input values to the nearest quantization point.
However, these methods are highly sensitive to outliers, which can excessively expand the dynamic range of the target tensor. Such over-scaling often results in underflow, especially in extremely low-bit representation.
To solve this problem, we use the standard deviation of the input tensor as a scaling factor. Unlike the \textit{absmax} value, the standard deviation is more robust to outliers. Moreover, since our DANUQ is based on a probability model of a standard normal distribution, the standard deviation provides a more reliable approximation of the target dynamic range.

In FedWSQ, we use a shared global scaling factor to help maintain consistency in quantization. Instead of independently using scaling factors, we employ a collaborative approach in which each client contributes to a global scaling vector, ensuring that local standard deviations are taken into account.
Specifically, each client $i$ computes a scaling vector, defined as $\mathbf{s}_{i} = [s_{i,1},\dots,s_{i,L}]^T,$
where $s_{i,l}$ denotes the standard deviation of the LMPUs for the $l$-th layer.
The server then collects and aggregates $\mathbf{s}_i$ from the participating clients $i \in \mathcal{S}$, and updates the global scaling vector $\mathbf{s}_\textrm{g} = [s_{\textrm{g},1},\dots,s_{\textrm{g},L}]^T$ as follows:
\begin{equation}
    \mathbf{s}_\textrm{g} \leftarrow (1-\beta) \mathbf{s}_\textrm{g} + \beta \frac{1}{|\mathcal{S}|} \sum_{i \in \mathcal{S}} \mathbf{s}_i,
    \label{eq:global_scaler}
\end{equation}
where $\beta$ is a momentum parameter that controls the update rate. This approach ensures that the global scaling factor smoothly adapts while mitigating local fluctuations. Once updated, the server distributes $\mathbf{s}_\textrm{g}$ to the participating clients.
Each client then divides its LMPU updates for the $l$-th layer by the corresponding scale $s_{\mathrm{g},l}$, so that the normalized values can be regarded as samples from a standard normal distribution. Based on this assumption, we apply the proposed DANUQ scheme detailed in the next section.

% Each client then uses $s_{\textrm{g},l}$ to normalize its LMPUs for the $l$-th layer during quantization, following Eq.~(\ref{eq:quant_func}).

\vspace{-3mm}
\paragraph{QLs considering normal distribution}
To determine the optimal multi-bit QLs, we formulate an optimization problem that minimizes the expected quantization error under a normal distribution. We assume that a random variable $\Delta w$ from the normalized LMPUs follows a standard normal distribution.
Since the distribution is symmetric around zero, we consider only its nonnegative half.
Given a $B$-bit representation, let $Q=\lbrace q_0,q_1,\dots,q_R\rbrace$ be the set of QLs, where $R=2^{B-1}-1$. Here, we assume that $q_r$ values are sorted in ascending order and set $q_0=0$ as a fixed value.
Then, the quantization boundaries are defined as
\begin{equation}
u_r =
\begin{cases} 
q_0, & \text{if } r = 0, \\ 
\frac{q_{r-1} + q_{r}}{2}, & \text{if } 1 \leq r \leq R, \\ 
+\infty, & \text{if } r = R + 1.
\end{cases}
\label{eq:edges}
\end{equation}
A full-precision value in the range $[u_r, u_{r+1})$ is quantized into the QL $q_r$. The optimal QLs are determined by minimizing the expected quantization error, formulated as
\begin{equation}
\mathbb{E} \left[ (\Delta w - \Delta \bar{w} \right)^2] = 
\sum_{r=0}^{R} \int_{u_r}^{u_{r+1}} (x - q_r)^2 p(x) dx,
\label{eq:optim}
\end{equation}
where $\Delta \bar{w} \in Q$ represents the quantized value, and $p(x)$ is the PDF of the standard normal distribution. From~(\ref{eq:optim}), we can derive the following optimization problem as
\begin{align}
\min_{\{q_1,\dots,q_R\}} & \bigg\{ \frac{1}{2}(q_{R}^2+1) 
-\sqrt{\frac{2}{\pi}} q_1 e^{-\frac{u_1^2}{2}}
-\frac{1}{2} q_1^2 \textrm{erf}\left( \frac{u_1}{\sqrt{2}} \right) \nonumber\\
& + \sqrt{\frac{2}{\pi}} \sum_{r=1}^{R-1} (q_r-q_{r+1}) e^{-\frac{u_{r+1}^2}{2}} \nonumber \\
& + \frac{1}{2} \sum_{r=1}^{R-1} (q_r^2-q_{r+1}^2)\textrm{erf}\left( \frac{u_{r+1}}{\sqrt{2}} \right)
\bigg\},
\label{eq:cost_func}
\end{align}
where $\mathrm{erf}(\cdot)$ denotes the error function.\footnote{The full derivation is provided in the supplementary document.} 
A closed-form solution to Eq.~(\ref{eq:cost_func}) is intractable due to the presence of nonlinear terms, including the Gaussian integral and the error function, which lack simple analytical inverses, and the interdependent QLs due to boundary conditions. 
As an alternative, we employ a brute-force search algorithm to numerically determine the optimal QLs.\footnote{We discretize the search space of possible QLs and perform an exhaustive evaluation of the cost function to identify the configuration that minimizes the quantization error.
To improve computational efficiency, we restrict the search space to a reasonable range based on empirical observations and employ parallel processing to accelerate evaluation.}
The optimal QLs for different bit-widths are obtained as follows: \textbf{1) 1-bit}\footnote{For the 1-bit case, the QLs consist of only two values. Thus, the constraint, $q_0 = 0$, is omitted to allow optimal placement of both QLs.}: \mbox{[-0.798, 0.798]}; \textbf{2) 2-bit}: [-1.224, 0, 0.765, 1.724]; \textbf{3) 4-bit}: [-2.654, -1.974, -1.508, -1.149, -0.834, -0.544, -0.269, 0, 0.230, 0.465, 0.708, 0.966, 1.248, 1.568, 1.968, 2.649].
These QLs minimize quantization error under the normal distribution, which provides an effective trade-off between precision and computational efficiency.

\begin{algorithm}[t]
\footnotesize
\caption{FedWSQ}
\label{alg:fedwsq}
\SetAlgoLined
\KwIn{Initial global parameters $\mathbf{W}_\textrm{g}^0$, initial scales $\mathbf{s}_\textrm{g}^0$, global communication round $T$, local training iteration $K$, number of total clients $N$, number of model layers $L$, learning rate $\eta$, hyperparameter $\beta$}
% \KwIn{Initial global parameters $\mathbf{W}_\textrm{g}^0$, initial scales $\mathbf{s}_\textrm{g}^0$, global communication round $T$, local training iteration $K$, number of total clients $N$, number of model layers $L$, learning rate $\eta$}

\textbf{Server-side:}

\For{$t = 1, \dots ,T$}{
    Randomly sample a subset of clients $\mathcal{S}_t \subseteq [N]$

    Set the bit-precision $B_i$ for the client $i \in \mathcal{S}_t$
    
    Send $(\mathbf{W}_\textrm{g}^{t-1}, \mathbf{s}_\textrm{g}^{t-1})$ to client $ i \in \mathcal{S}_t$

    \For{$\mathrm{each~ client}~ i \in \mathcal{S}_t,~\textbf{in parallel}$}{
        $( \Delta \overline{\mathbf{W}}_i^t, \mathbf{s}_i^t) \leftarrow \textbf{ClientLocalTraining} ( \mathbf{W}_\textrm{g}^{t-1}, \mathbf{s}_\textrm{g}^{t-1}, i, B_i)$
    }
    \For{$i \in \mathcal{S}_t$}{
        Obtain dequantized parameters $\Delta_i^t$ using $( \Delta \overline{\mathbf{W}}_i^t, \mathbf{s}_i^t)$
    }

    Aggregate local updates: $\Delta^t \leftarrow \sum_{i \in \mathcal{S}_t} \Delta_i^t$

    Update global parameters: $\mathbf{W}_\textrm{g}^t \leftarrow \mathbf{W}_\textrm{g}^{t-1} + \Delta^t$

    Update global scales: $\mathbf{s}_\textrm{g}^t \leftarrow (1-\beta) \mathbf{s}_\textrm{g}^{t-1} + \beta \frac{1}{|\mathcal{S}_t|} \sum_{i \in \mathcal{S}_t} \mathbf{s}_i^t$
    % Update global scales $\mathbf{s}_\textrm{g}^t$ using (\ref{eq:global_scaler})

    }

\hrulefill

\textbf{ClientLocalTraining}$(\mathbf{W}_\textrm{g}, \mathbf{s}_\textrm{g}, i, B)$:

Initialize local parameters: $\mathbf{W}_i^{0} \leftarrow \mathbf{W}_\textrm{g}$

Set DANUQ function $\mathcal{Q}(\cdot)$ based on the bit-precision $B$

\For{$k = 1, \dots ,K$}{
    Apply WS to $\mathbf{W}_i^{k-1}$ using (\ref{eq:ws})

    Compute unbiased gradient of local loss $\nabla f_i(\mathbf{W}_i^{k-1})$

    Update local parameters: $\mathbf{W}_i^{k} \leftarrow \mathbf{W}_i^{k-1} - \eta \nabla f_i(\mathbf{W}_i^{k-1})$
}

Compute local update: $\Delta \mathbf{W}_i \leftarrow \mathbf{W}_i^{K} - \mathbf{W}_\textrm{g}$

\For{$\mathrm{each~ layer}~$ $l = 1, \dots ,L$}{
    Compute quantized local update:
    $\Delta \overline{\mathbf{W}}_{i,l} \leftarrow \mathcal{Q}(\Delta \mathbf{W}_{i,l} / s_{\textrm{g},l})$

    Calculate local scale factors: $s_{i,l} \leftarrow \textrm{std}(\Delta \mathbf{W}_{i,l})$
}

$\Delta \overline{\mathbf{W}}_i\coloneq(\Delta \mathbf{W}_{i,1},\cdots,\Delta \mathbf{W}_{i,L})$

$\mathbf{s}_i\coloneq [s_{i,1},\cdots,s_{i,L}]$

\KwRet{$(\Delta \overline{\mathbf{W}}_i, \mathbf{s}_i)$}
\end{algorithm}

\begin{table*}[t]
\begin{center}
\caption{FL performance on the three benchmark datasets with a 5\% participation rate over 100 clients for $\alpha \in \{0.1, 0.3, 0.6\}$.}
\vspace{-2mm}
% The arrows indicate whether higher ($\uparrow$) or lower ($\downarrow$) is better.
% FedCM$^\dagger$ and FedDC$^\ddagger$ require 50\% and 100\% additional communication costs for each communication round, respectively.}
\label{tab:main1}
% \vspace{-0.2cm}
\setlength\tabcolsep{7.9pt}
\hspace{-0.2cm}
\scalebox{0.72}{
\begin{tabular}{lccccccccccccc} 
%\cline{1-13}
\toprule
\multirow{2}{*}{Method} & \multirow{2}{*}{\#bits}
 & \multicolumn{4}{c}{CIFAR-10} & \multicolumn{4}{c}{CIFAR-100} & \multicolumn{4}{c}{Tiny-ImageNet} \\
 \cmidrule(lr){3-6} \cmidrule(lr){7-10} \cmidrule(lr){11-14}
& & $\alpha = 0.1$ & 0.3 & 0.6 & \textit{i.i.d.} & 0.1 & 0.3 & 0.6 & \textit{i.i.d.} & 0.1 & 0.3 & 0.6 & \textit{i.i.d.} \\
%\cline{1-9}
\midrule
FedAvg~\cite{mcmahan2017communication} & 32 & 70.65 & 82.52 & 85.45 & 89.25 & 43.45 & 47.39 & 49.04 & 48.66 & 30.37 & 34.65 & 36.54 & 37.39 \\
FedProx~\cite{li2020federated} & 32 & 70.89 & 83.91 & 86.60 & 88.81 & 43.78 & 48.26 & 48.49 & 48.54 & 29.99 & 35.44 & 37.14 & 37.43 \\
FedAvgM~\cite{hsu2019measuring} & 32 & 78.30 & 85.27 & 87.94 & 90.00 & 47.70 & 52.04 & 53.00 & 53.40 & 32.00 & 37.63 & 39.66 & 41.02 \\
FedADAM~\cite{reddi2021adaptive} & 32 & 72.58 & 81.73 & 85.97 & 88.45 & 46.83 & 53.43 & 55.96 & 57.93 & 34.77 & 39.97 & 43.27 & 46.26 \\
FedDyn~\cite{acar2021federated} & 32 & 83.63 & 88.28 & 89.20 & 91.08 & 53.83 & 56.40 & 56.90 & 56.93 & 37.32 & 40.92 & 42.78 & 42.65 \\
FedMLB~\cite{kim2022multi} & 32 & 71.12 & 82.40 & 86.23 & 89.25 & 49.72 & 54.16 & 55.17 & 56.45 & 33.45 & 39.06 & 40.55 & 41.46 \\
FedLC~\citep{zhang2022Federated} & 32 & 74.03 & 83.80 & 86.20 & 88.35 & 43.86 & 48.50 & 48.61 & 47.81 & 31.73 & 34.55 & 36.72 & 38.06 \\
FedNTD~\cite{lee2022preservation} & 32 & 73.29 &  83.34 & 86.26 & 89.36 & 45.08 & 48.73 & 50.39 & 50.90 & 34.36 & 36.44 & 39.06 & 40.60 \\
FedSmoo~\cite{sun2023dynamic} & 32 & 78.61 & 80.07 & 85.03 & 86.48 & 42.02 & 44.97 & 45.32 & 49.43 & 23.09 & 27.14 & 29.70 & 34.96 \\
FedDecorr~\citep{shi2022towards} & 32 & 75.44 & 83.01 & 84.49 & 88.40 & 44.05 & 49.35 & 50.44 & 49.40 & 31.12 & 34.13 & 35.59 & 36.36 \\
FedWon~\citep{zhuang2024fedwon} & 32 & 70.89 & 78.36 & 82.44 & 86.56 & 45.55 & 51.68 & 53.60 & 55.28 & 30.52 & 36.20 & 38.59 & 40.73 \\
FedRCL~\cite{seo2024relaxed} & 32 & 83.45 & 88.19 & 89.74 & 91.55 & 58.26 & 62.21 & 63.95 & 65.11 & 37.86 & 44.94 & 46.75 & 47.96 \\
FedACG~\cite{kim2024fedacg} & 32 & 83.62 & 89.12 & 90.38 & 91.75 & 58.14 & 62.80 & 61.85 & 62.46 & 39.75 & 45.46 & 48.22 & 50.49\\
\midrule
\multirow{2}{*}{FedPAQ~\cite{reisizadeh2020fedpaq}} & 4 & 66.81 & 78.52 & 81.36 & 85.46 & 38.06 & 42.42 & 43.46 & 44.35 & 28.54 & 31.93 & 33.56 & 34.98 \\
& 1 & 57.82 & 71.27 & 72.36 & 79.46 & 30.07 & 34.16 & 38.24 & 37.85 & 24.74 & 29.48 & 30.74 & 33.81 \\
\cmidrule{2-14}
\multirow{2}{*}{FedHQ+~\cite{Chen2021fedHQ+}} & 4 & 66.89 & 78.42 & 81.42 & 85.54 & 38.22 & 42.66 & 43.10 & 44.86 & 28.59 & 32.18 & 34.24 & 35.69 \\
& 1 & 58.14 & 70.68 & 73.16 & 79.04 & 31.11 & 34.11 & 37.50 & 39.17 & 24.79 & 29.18 & 31.14 & 33.50 \\
\midrule
\multirow{5}{*}{FedWSQ} & FedWS~(32) & \cellcolor{Gray} 85.94 & \underline{89.81} & \underline{91.00} & \textbf{92.47} & \underline{64.14} & \textbf{68.19} & \textbf{69.31} & \textbf{70.14} & \textbf{47.05} & \textbf{52.84} & \textbf{54.23} & \textbf{55.26} \\
& 4 & \textbf{86.84} & \textbf{89.84} & \textbf{91.11} & \underline{92.21} & \textbf{64.34} & \underline{67.31} & \underline{68.41} & \underline{68.53} & \cellcolor{Gray} 46.51 & \cellcolor{Gray} 50.49 & \underline{51.96} & \underline{52.13} \\
& 1 & 84.79 & 88.62 & 89.49 & 91.15 & 62.05 & 65.14 & 66.19 & 66.11 & 45.11 & 49.93 & 51.11 & 51.77 \\
& FBA~(2.33) & \underline{86.16} & \cellcolor{Gray} 89.60 & \cellcolor{Gray} 90.44 & 91.71 & 63.07 & 66.16 & 67.08 & 67.17 & \underline{46.62} & \underline{50.69} & 51.43 & 51.68 \\
& DBA~(2.33) & 85.20 & 89.23 & 90.26 & \cellcolor{Gray} 91.78 & \cellcolor{Gray} 63.28 & \cellcolor{Gray} 66.32 & \cellcolor{Gray} 67.26 & \cellcolor{Gray} 67.22 & 46.50 & 50.24 & \cellcolor{Gray} 51.55 & \cellcolor{Gray} 51.80  \\
\bottomrule
\end{tabular}}
\end{center}
\vspace{-7mm} 
\end{table*}

\vspace{-2mm}
\paragraph{Mixed channel-bit allocation}
To further enhance communication efficiency in FL, we propose an adaptive mixed-precision strategy in which each client dynamically selects its bit representation based on its communication channel conditions. This approach effectively balances model update quality and transmission cost, reducing overall bandwidth utilization. 
To validate this strategy, we conduct two simulation setups: 1) Fixed-bit allocation~(FBA), where each client keeps a constant bit-width selected from $\{1, 2, 4\}$ throughout training, and 2) dynamic-bit allocation~(DBA), where each client is assigned a randomly selected bit-width at every communication round. 
With bit-width selection following a uniform distribution over $\{1,2,4\}$, the expected bit-width per client is approximately 2.3 bits.
The entire FedWSQ framework, including the mixed-precision strategy, is summarized in Algorithm~\ref{alg:fedwsq}.

\begin{table*}[t]
\begin{center}
% \vspace{-3mm}
\caption{Comparison of UQ and DANUQ methods with or without WS on the three benchmark datasets with a 5\% participation rate over 100 clients for $\alpha \in \{0.1, 0.3, 0.6\}$.}
\vspace{-2mm}
\label{tab:main2}
% \vspace{-0.2cm}
\setlength\tabcolsep{7.9pt}
\hspace{-0.2cm}
\scalebox{0.69}{
\begin{tabular}{ccccccccccc} 
%\cline{1-13}
\toprule
\multirow{2}{*}{Dataset} & \multirow{2}{*}{WS} & \multirow{2}{*}{\#bits} & \multicolumn{2}{c}{$\alpha=0.1$} & \multicolumn{2}{c}{$\alpha=0.3$} & \multicolumn{2}{c}{$\alpha=0.6$} & \multicolumn{2}{c}{\textit{i.i.d.}} \\
\cmidrule(lr){4-5} \cmidrule(lr){6-7} \cmidrule(lr){8-9} \cmidrule(lr){10-11}
 & & & UQ & DANUQ & UQ & DANUQ & UQ & DANUQ & UQ & DANUQ \\ 
%\cline{1-9}
\midrule
\multirow{6}{*}{CIFAR-10} & \multirow{3}{*}{\ding{56}} & 1 & 57.82 & \textbf{72.48} & 71.27 & \textbf{80.88} & 72.36 & \textbf{83.52} & 79.46 & \textbf{86.92} \\
&  & 2 & 62.39 & \textbf{65.19} & 74.35 & \textbf{78.61} & 76.79 & \textbf{82.59} & 82.25 & \textbf{86.11} \\
&  & 4 & 66.81 & \textbf{72.87} & 78.52 & \textbf{84.62} & 81.36 & \textbf{86.26} & 85.46 & \textbf{88.66} \\
\cmidrule(lr){2-11}
& \multirow{3}{*}{\ding{52}} & 1 & 75.48 & \textbf{84.79} & 80.15 & \textbf{88.62} & 82.53 & \textbf{89.49} & 85.37 & \textbf{91.15} \\
&  & 2 & 78.36 & \textbf{85.92} & 83.02 & \textbf{88.98} & 84.10 & \textbf{90.35} & 86.95 & \textbf{91.50} \\
&  & 4 & 82.78 & \textbf{86.84} & 86.37 & \textbf{89.84} & 87.58 & \textbf{91.11} & 89.87 & \textbf{92.21} \\
\midrule
\multirow{6}{*}{CIFAR-100} & \multirow{3}{*}{\ding{56}} & 1 & 30.07 & \textbf{40.51} & 34.16 & \textbf{46.62} & 38.24 & \textbf{48.67} & 37.85 & \textbf{48.55} \\
&  & 2 & 33.52 & \textbf{40.58} & 37.66 & \textbf{45.07} & 40.47 & \textbf{46.31} & 40.90 & \textbf{46.06} \\
&  & 4 & 38.06 & \textbf{43.59} & 42.42 & \textbf{48.33} & 43.46 & \textbf{49.29} & 44.35 & \textbf{48.76} \\
\cmidrule(lr){2-11}
& \multirow{3}{*}{\ding{52}} & 1 & 45.59 & \textbf{62.05} & 50.10 & \textbf{65.14} & 52.13 & \textbf{66.19} & 54.58 & \textbf{66.11} \\
&  & 2 & 49.68 & \textbf{63.34} & 54.22 & \textbf{65.96} & 56.01 & \textbf{67.19} & 58.72 & \textbf{67.82} \\
&  & 4 & 56.85 & \textbf{64.34} & 60.21 & \textbf{67.31} & 61.98 & \textbf{68.41} & 62.98 & \textbf{68.53} \\
\midrule
\multirow{6}{*}{Tiny-ImageNet} & \multirow{3}{*}{\ding{56}} & 1 & 24.74 & \textbf{28.95} & 29.48 & \textbf{33.14} & 30.74 & \textbf{36.75} & 33.81 & \textbf{38.95} \\
&  & 2 & 27.02 & \textbf{30.33} & 31.60 & \textbf{34.12} & 33.58 & \textbf{36.6} & 34.87 & \textbf{35.41} \\
&  & 4 & 28.54 & \textbf{31.14} & 31.93 & \textbf{35.42} & 33.56 & \textbf{36.66} & 34.98 & \textbf{38.13} \\
\cmidrule(lr){2-11}
& \multirow{3}{*}{\ding{52}} & 1 & 32.73 & \textbf{45.11} & 37.63 & \textbf{49.93} & 39.60 & \textbf{51.11} & 41.98 & \textbf{51.77} \\
&  & 2 & 36.16 & \textbf{45.96} & 40.90 & \textbf{50.17} & 42.51 & \textbf{51.12} & 44.89 & \textbf{51.12} \\
&  & 4 & 40.46 & \textbf{46.51} & 44.58 & \textbf{50.49} & 45.55 & \textbf{51.96} & 46.93 & \textbf{52.13} \\
\bottomrule
\end{tabular}}
\end{center}
\vspace{-7mm}
\end{table*}

\begin{table}[t]
\begin{center}
\caption{Comparison of NUQ methods using WS on CIFAR-10 and CIFAR-100 datasets for $\alpha \in \{0.1, 0.3, 0.6\}$.}
\vspace{-0.2cm}
\label{tab:uniform_quant}
% \vspace{-0.2cm}
\setlength\tabcolsep{7.9pt}
\hspace{-0.2cm}
\scalebox{0.68}{
\begin{tabular}{clcccccc} 
%\cline{1-13}
\toprule
\multirow{2}{*}{\#bits} & \multirow{2}{*}{Method} & \multicolumn{3}{c}{CIFAR-10} & \multicolumn{3}{c}{CIFAR-100} \\
\cmidrule(lr){3-5} \cmidrule(lr){6-8}
& & $\alpha=0.1$ & 0.3 & 0.6 & 0.1 & 0.3 & 0.6 \\ 
%\cline{1-9}
\midrule
\multirow{2}{*}{1} & NF/FP & \underline{24.0} & \underline{30.6} & \underline{28.8} & \underline{7.0} & \underline{8.3} & \underline{8.1} \\
& DANUQ & \textbf{84.8} & \textbf{88.6} & \textbf{89.5} & \textbf{62.1} & \textbf{65.1} & \textbf{66.2} \\
\midrule
\multirow{3}{*}{2} & NF & \underline{57.8} & \underline{73.8} & \underline{80.0} & \underline{41.2} & \underline{49.1} & \underline{53.0} \\
& FP & 45.0 & 57.5 & 58.6 & 25.3 & 32.0 & 34.2 \\
& DANUQ & \textbf{85.9} & \textbf{89.0} & \textbf{90.4} & \textbf{63.3} & \textbf{66.0} & \textbf{67.2} \\
\midrule
\multirow{3}{*}{4} & NF & 86.6 & 89.7 & \underline{91.0} & \textbf{64.8} & 66.7 & \textbf{68.4} \\
& FP & \textbf{87.1} & \textbf{90.0} & \underline{91.0} & \underline{64.3} & \underline{67.0} & \textbf{68.4} \\
& DANUQ & \underline{86.8} & \underline{89.8} & \textbf{91.1} & \underline{64.3} & \textbf{67.3} & \textbf{68.4} \\
\bottomrule
\end{tabular}}
\end{center}
\vspace{-7mm}
\end{table}

% !TEX root = ../main.tex

\section{Experiments}
\label{sec:exp} 
%This section presents an evaluation of the proposed WSQ method in various FL scenarios. We compare WSQ-enhanced models with state-of-the-art FL algorithms using \textit{i.i.d.} and non-\textit{i.i.d.} data, where each device utilizes either a fixed bit-width or an adaptive bit-width strategy. We also analyze the impact of hyper-parameters and offer further analysis of the proposed method. The convergence plots obtained from the experiments are depicted in Section~\ref{ssup:convergence_plot} of the supplementary document.

In this section, we evaluate the proposed FedWSQ on standard FL benchmarks, compare its performance against various FL methods on both \textit{i.i.d.} and non-\textit{i.i.d.} data conditions, and offer a further analysis of FedWSQ. The convergence plots obtained from the experiments are provided in Section~\ref{ssup:convergence_plot} of the supplementary document.

%%%%%%%%%%%%%%%%%%%%%%%%% Main result %%%%%%%%%%%%%%%%%%%%%%%

\subsection{Experimental setup}
\label{sec:setup}
For experiments, we use three standard benchmark datasets: CIFAR-10~\citep{krizhevsky2009learning}, CIFAR-100~\citep{krizhevsky2009learning}, and Tiny-ImageNet~\citep{le2015tiny}.
CIFAR-10 and CIFAR-100 contain 60,000 images divided into 10 and 100 classes, respectively. Tiny-ImageNet is composed of 200 classes, providing a more complex image classification task.
For \textit{i.i.d.} settings, training samples are randomly assigned to clients without replacement.
For the non-\textit{i.i.d.} setting, we sample label ratios from a Dirichlet distribution with a concentration parameter $\alpha \in \{0.1, 0.3, 0.6\}$, following Hsu~\etal~\cite{hsu2019measuring}.
Lower values of $\alpha$ (\textit{e.g.}, 0.1) indicate higher data heterogeneity, whereas larger values (\textit{e.g.}, 0.6) and the \textit{i.i.d.} case correspond to more homogeneous distributions.
The participation rate is set to $5\%$ among 100 clients.
Model accuracy is evaluated on the test set of each dataset at the 1,000th communication round.
To ensure stable evaluations, we report accuracy using an exponential moving average with a smoothing parameter of 0.9.
To validate the effectiveness of FedWSQ, we integrate it with FedAvg~\cite{mcmahan2017communication}. Note that FedWSQ can be incorporated into any existing FL algorithm.
For clarity, we refer to the proposed method without DANUQ as FedWS.
For quantized FL settings, we consider three bit-allocation strategies: constant allocation, FBA, and DBA. In the constant allocation settings, a fixed bit-width is assigned to all clients.
We compare FedWSQ against various FL algorithms, including FedAvg~\citep{mcmahan2017communication}, FedProx~\citep{li2020federated}, FedAvgM~\citep{hsu2019measuring}, FedADAM~\citep{reddi2021adaptive}, FedDyn~\citep{acar2021federated},
FedMLB~\cite{kim2022multi}, FedLC~\cite{zhang2022Federated},
FedNTD~\cite{lee2022preservation}, FedSmoo~\cite{sun2023dynamic}, FedDecorr~\cite{shi2022towards}, FedWon~\citep{zhuang2024fedwon}, FedRCL~\cite{seo2024relaxed}, FedACG~\cite{kim2024fedacg}, FedPAQ~\cite{reisizadeh2020fedpaq}, and FedHQ+~\cite{Chen2021fedHQ+}.

\subsection{Implementation details}
We follow most of the implementation setups and evaluation protocols from~\cite{kim2024fedacg,acar2021federated,xu2021fedcm,seo2024relaxed}.
The ResNet-18 architecture~\citep{he2016deep} is adopted as the backbone network. Consistent with standard practice in FL~\cite{hsieh2020non}, all BN~\cite{ioffe2015batch} layers in ResNet-18 are replaced with GN~\cite{wu2018group} layers. 
Following the recommendation of \cite{qiao2019ws}, WS is applied before each GN layer to achieve optimal performance.
%This also follows the recommendation of Qiao~\etal~\cite{qiao2019ws} to use WS in combination with BN or GN layers for optimal performance. 
The hyper-parameter of WS is set to $\rho=0.001$, and the momentum for updating the scaling vector is set to $\beta=0.1$.
Additional information on the hyper-parameter selection is provided in Section~\ref{supp:exp_setup} of the supplementary document.

\subsection{Comparison with existing FL methods}
\label{ssec:sota}
%\paragraph{Evaluation in standard settings}

Table~\ref{tab:main1} presents the performance comparison of the proposed FedWSQ against various existing FL methods on the three benchmark datasets.
The bit-width settings for each FL method are indicated in the table.
Compared to FL methods that use quantization, such as FedPAQ and FedHQ+, FedWSQ demonstrates superior performance for 4-bit and 1-bit settings.
Also, FedWSQ outperforms FL methods with full-precision representation, including SOTA methods such as FedACG and FedRCL.
Notably, even with a 1-bit representation, FedWSQ demonstrates competitive performance on CIFAR-10 and achieves superior results on CIFAR-100 and Tiny-ImageNet.
For example, in a highly heterogeneous setting~($\alpha=0.1$), FedWSQ~(1-bit) achieves 62.05\% accuracy on CIFAR-100, surpassing FedRCL and FedACG by 3.79\% and 3.91\%, respectively. Similarly, on Tiny-ImageNet, FedWSQ~(1-bit) achieves 45.11\% accuracy, outperforming FedRCL and FedACG by 7.25\% and 5.36\%, respectively.
These results highlight the effectiveness of FedWSQ, which demonstrates its robustness even in extremely low bit-width scenarios while maintaining high model accuracy.

\begin{figure}[t]
\centering
\includegraphics[width=\linewidth]{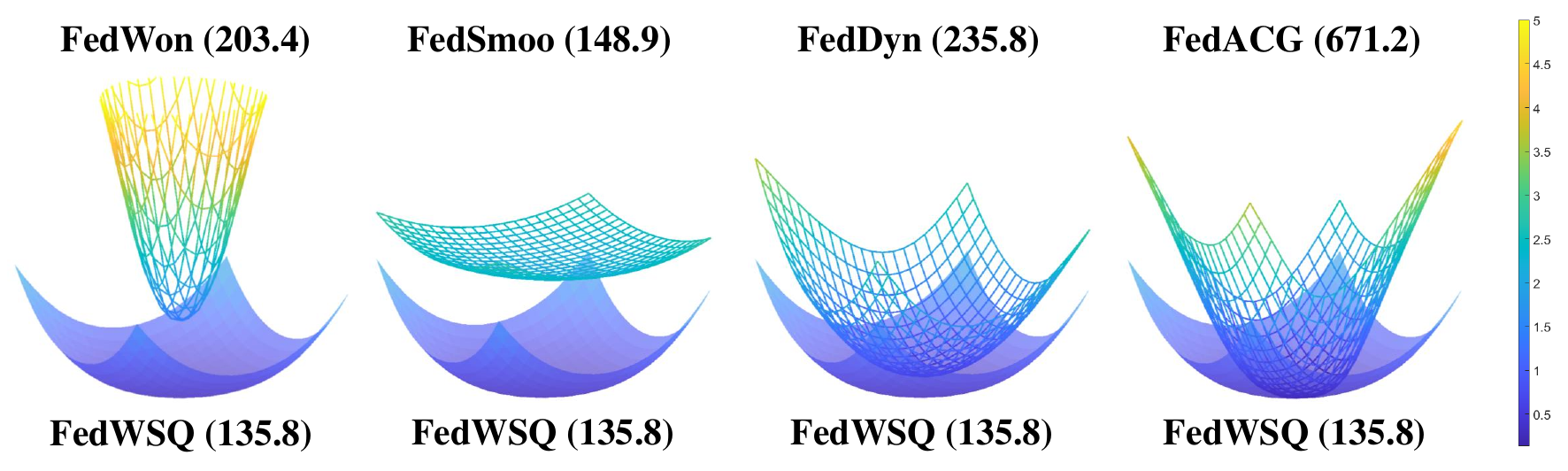}
\vspace{-0.6cm}
\caption{Loss landscape of FL methods trained on CIFAR-100, where the numbers indicate the Hessian top eigenvalue.
}
\label{fig:loss_landscape}
\vspace{-0.5cm}
\end{figure}

\subsection{Additional analyses}
\label{ssec:analysis}
\paragraph{Impact of bit representation in FedWSQ}
\label{ssec:bit}
FedWSQ balances model performance and communication efficiency by adopting different bit-width settings. Table~\ref{tab:main1} presents the results for 32-bit~(FedWS), 4-bit, and 1-bit representations, as well as adaptive bit-allocation strategies (FBA and DBA).
Despite using lower bit-widths, FedWSQ remains robust in non-\textit{i.i.d.} settings, particularly under high heterogeneity~($\alpha=0.1$).
Since local models tend to overfit under high data heterogeneity, quantizing model updates helps to regularize overfitting, achieving more robust FL performance~\cite{askarihemmat2022qreg,askarihemmat2024qgen}.
The accuracy gaps among different bit-widths are relatively small, and thus the FBA and DBA strategies enable efficient communication with marginal performance loss.
As shown in Table~\ref{tab:main1}, both FBA and DBA strategies achieve performance close to the 4-bit setting while reducing communication costs by half.
In summary, FedWSQ confirms its ability to maintain accuracy while significantly reducing communication bandwidth usage for real-world FL scenarios.

\vspace{-3mm}
\paragraph{Effect of WS and DANUQ}
Table~\ref{tab:main2} shows a comparative analysis of our proposed DANUQ and UQ under varying bit-widths and data heterogeneity.
In this experiment, UQ is implemented using the FedPAQ approach~\cite{reisizadeh2020fedpaq}.
As listed in the table, our DANUQ consistently outperforms the UQ method. 
Under extremely low-bit quantization (1-bit and 2-bit settings), the UQ suffers from significant accuracy degradation. 
While WS alone improves performance, combining WS with the proposed DANUQ yields the highest performance gains. 
The UQ with WS shows improvements compared with its non-WS counterparts, but the synergy between WS and DANUQ provides the most robust FL performance. 
Furthermore, our method maintains high accuracy even with low-bit representations. 
This suggests that applying FBA or DBA can be highly effective, as our approach maintains strong performance across varying bit-widths and makes FedWSQ particularly practical for real-world FL deployments.

\vspace{-4mm}
\paragraph{Comparison with other NUQ methods}
We evaluate the effectiveness of the proposed DANUQ by comparing it with existing NUQ approaches, FP~\cite{wang2025optimizing} and NF~\cite{dettmers20228bit}.\footnote{QLs of different methods are visualized in Section~\ref{supp:exp_setup} of the supplementary document.} As shown in Table~\ref{tab:uniform_quant}, at the 4-bit representation, all NUQ methods perform competitively, indicating that NUQ is generally robust at relatively higher bit-widths.
However, under 1-bit and 2-bit representations, FP and NF suffer from severe performance degradation, particularly in highly heterogeneous data~($\alpha=0.1$). In contrast, our DANUQ outperforms both NF and FP across all conditions, which confirms its robustness against extreme quantization constraints.
These results highlight the superiority of our DANUQ method among NUQ methods for communication-efficient FL.

\begin{table}[t]
    \centering
    \caption{FL performance of different backbone architectures for $\alpha=0.3$.
     }
    \vspace{-0.2cm}
    \scalebox{0.7}{
    \setlength\tabcolsep{10pt}
    \hspace{-0.25cm}
    \begin{tabular}{lccc}
        \toprule
        Backbone & Vanilla & FedWS & FedWSQ (DBA)  \\
        \midrule
        ShuffleNet~\cite{zhang2018shufflenet} & 36.37 & \textbf{53.32} & \underline{51.12} \\
        VGGNet-9~\cite{simonyan2014vgg} & 46.14 & \underline{60.01} & \textbf{60.48} \\
        SqueezeNet~\cite{iandola2016squeeze} & 39.45 & \underline{55.87} & \textbf{56.10} \\
        ResNet-18~\cite{he2016deep} & 47.39 & \textbf{68.19} & \underline{66.32} \\
        MobileViT~\cite{mehta2021mobilevit} & 35.42 & \underline{40.80} & \textbf{41.27} \\
        \bottomrule
    \end{tabular}}
    \label{tab:backbone}
    \vspace{-5mm}
\end{table}

\vspace{-4mm}
\paragraph{Loss Landscape Analysis}
Figure \ref{fig:loss_landscape} visualizes the loss landscapes of various FL methods compared to FedWSQ~(with DBA). FedWSQ achieves the smoothest and most stable loss landscape, with the lowest Hessian top eigenvalue of 135.8, indicating improved generalization ability. FedSmoo exhibits a flatter landscape but converges to a much higher global minimum, reducing sharpness at the cost of performance. 
In contrast, FedACG reaches a global comparable to FedWSQ but has a significantly sharper landscape, with a Hessian top eigenvalue of 671.2, which indicates it is more sensitive to perturbations.
These results show that FedWSQ effectively enhances the generalization, leading to more stable training and improved FL performance.

\vspace{-3mm}
\paragraph{Impact on different backbone architectures}
Table~\ref{tab:backbone} presents the performance of different backbone architectures on CIFAR-100 to evaluate the impact of FedWS and FedWSQ. 
For all architectures, FedWS significantly improves accuracy over the vanilla models, and FedWSQ maintains a comparable performance to FedWS with improved communication efficiency.
In addition, We evaluate MobileViT~\cite{mehta2021mobilevit}, a vision transformer-based architecture, which also benefits from FedWS and FedWSQ. These results confirm their applicability beyond CNNs.

\begin{table}[t]
    \centering
    \caption{Ablation study for $\rho$ using FedWS on CIFAR-10. For the non-\textit{i.i.d.} setting, $\alpha=0.3$ is used.
    }
    \vspace{-0.2cm}
    \scalebox{0.7}{
    \setlength\tabcolsep{10pt}
    \hspace{-0.25cm}
    \begin{tabular}{ccccc}
        \toprule
        $\rho$ & $1 \times 10^{-4} $ & $1 \times 10^{-3} $ & $1 \times 10^{-2} $ & $1 \times 10^{-1} $ \\
        \midrule
        non-\textit{i.i.d.} & 87.15 & \textbf{89.71} & 89.62 & 89.46 \\ 
        \textit{i.i.d.} & 90.64 & \textbf{92.48} & 91.97 & 92.11 \\ 
        \bottomrule
    \end{tabular}}
    \label{tab:rho}
    \vspace{-5mm}
\end{table}

\vspace{-4mm}
\paragraph{Comparision with FedWon}
We provide the comparison of FedWSQ with FedWon, which was discussed in Section~\ref{sec:comp_fedwon}.
As shown in Table~\ref{tab:main1}, FedWSQ substantially outperforms FedWon across all datasets and data heterogeneity settings. As illustrated in Figure~\ref{fig:loss_landscape}, FedWon exhibits a much higher global minimum and a very sharp loss landscape with a Hessian top eigenvalue of 203.4, whereas FedWSQ maintains a smoother surface with an eigenvalue of 135.8. These results demonstrate FedWSQ's superior robustness and generalization ability.

\vspace{-4mm}
\paragraph{Hyper-parameters}
We examine the hyper-parameter $\rho$ for WS, which handles the normalization scales. To evaluate its impact, we vary the values of $\rho$ and assess the performance of FedWS under both non-\textit{i.i.d.} and \textit{i.i.d.} data distributions. As shown in Table~\ref{tab:rho}, the performance of FedWS is insensitive to $\rho$ variations. This is because during inference, normalization layers like GN cancel the effect of constant scaling, and the variance of the parameter vectors within the same convolutional layer is not significant. We select $\rho= 1 \times 10^{-3}$, which yields the best results.
% !TEX root = ../main.tex

\section{Conclusions} 
\label{sec:conclusion}
This paper presents FedWSQ, a novel FL framework using WS and the proposed DANUQ to address key FL challenges, including data heterogeneity, partial client participation, and communication bottleneck. 
By leveraging WS, FedWSQ enhances training stability and mitigates client drift via an implicit gradient filtering mechanism. 
Additionally, DANUQ efficiently compresses LMPUs while preserving essential local information, significantly reducing communication overhead even at ultra-low bit-widths.
Extensive experiments on the benchmark datasets confirm that FedWSQ consistently outperforms SOTA FL methods under various conditions. Furthermore, the mixed-precision strategies FBA and DBA effectively balance communication cost and model accuracy.
These results validate the robustness of the FedWSQ framework in extreme quantization scenarios, making it highly practical for real-world FL.

% !TEX root = ../main.tex

\vspace{-0.2cm}
%\paragraph{Acknowledgments}
%\label{Acknowledgments}

{
    \small
    \bibliographystyle{ieeenat_fullname}
    \bibliography{main}
}

% !TEX root = ../main.tex
% CVPR 2024 Paper Template; see https://github.com/cvpr-org/author-kit

%\documentclass[10pt,twocolumn,letterpaper]{article}

%%%%%%%%% PAPER TYPE  - PLEASE UPDATE FOR FINAL VERSION
% \usepackage{cvpr}              % To produce the CAMERA-READY version
% To produce the REVIEW version
% \usepackage[pagenumbers]{cvpr} % To force page numbers, e.g. for an arXiv version

% Import additional packages in the preamble file, before hyperref
%\input{preamble}

% It is strongly recommended to use hyperref, especially for the review version.
% hyperref with option pagebackref eases the reviewers' job.
% Please disable hyperref *only* if you encounter grave issues, 
% e.g. with the file validation for the camera-ready version.
%
% If you comment hyperref and then uncomment it, you should delete *.aux before re-running LaTeX.
% (Or just hit 'q' on the first LaTeX run, let it finish, and you should be clear).
\newpage
\onecolumn
\definecolor{iccvblue}{rgb}{0.21,0.49,0.74}

\renewcommand\thesection{\Alph{section}}
\renewcommand\thetable{\Alph{table}}
\renewcommand\thefigure{\Alph{figure}}
\newcommand\numberthis{\addtocounter{equation}{1}\tag{\theequation}}
\appendix
\onecolumn

%%%%%%%%% PAPER ID  - PLEASE UPDATE
\def\paperID{9573} % *** Enter the Paper ID here
\def\confName{ICCV}
\def\confYear{2025}
\begin{center}
    \Large \textbf{FedWSQ: Efficient Federated Learning with Weight Standardization and Distribution-Aware Non-Uniform Quantization}\par
    \Large \textbf{\textit{--Supplementary Document--}}
    % \vspace{1em} % 제목과 본문 사이에 공간 추가
\end{center}

\begin{center}
    \large Paper ID 9573
    \vspace{1em} % 제목과 본문 사이에 공간 추가
\end{center}

\maketitle

\setcounter{table}{0}
\setcounter{figure}{0}
\setcounter{theorem}{0}
\setcounter{page}{1}
\setlength{\textfloatsep}{15pt}

\section{Technical Lemmas}
\label{supp:deriv1}
This section introduces some technical lemmas that are useful to understand our main document.

\begin{lemma}[]\label{lem:mean}
Consider any vector $\mathbf{v} \in \mathbb{R}^d$. The mean subtraction of $\mathbf{v}$ is given by
\begin{align*}
\bar{\mathbf{v}} &= \mathbf{v} - \left( \frac{1}{d} \mathbf{1}^T \mathbf{v} \right) \mathbf{1} \\
&= \left(\mathbf{I} - \frac{1}{d} \mathbf{1}\mathbf{1}^T \right) \mathbf{v} \\
&=\left(\mathbf{I} - \mathbf{P}_\mathbf{1} \right) \mathbf{v} \numberthis \label{eq:lemma1}
\end{align*}
where $\mathbf{I} \in \mathbb{R}^{d \times d}$ is the identity matrix, $\mathbf{1} \in \mathbb{R}^d$ is a vector whose elements are all ones, and $\mathbf{P}_\mathbf{w}$ represents the projection matrix onto the vector $\mathbf{w}$. Thus, mean subtraction is equivalent to projecting $\mathbf{v}$ onto $\operatorname{span}\lbrace \mathbf{1} \rbrace^\perp$. In other words, this projection removes the DC (constant) component from the given vector $\mathbf{v}$.
\end{lemma}

\begin{lemma}[]\label{lem:std}
Consider any vector $\bar{\mathbf{v}} \in \mathbb{R}^d$ with zero mean. Normalization of $\bar{\mathbf{v}}$ using its standard deviation $\sigma(\bar{\mathbf{v}})$ is given by
\begin{align*}
\tilde{\mathbf{v}} &= \frac{\rho}{\sigma(\bar{\mathbf{v}})}\bar{\mathbf{v}} \\
&= \frac{\rho\sqrt{d}}{\left\Vert \bar{\mathbf{v}} \right\Vert} \bar{\mathbf{v}}. \numberthis \label{eq:lemma2_1}
\end{align*}
Since $\bar{\mathbf{v}}$ is zero-centered, its standard deviation is given by $\sigma(\bar{\mathbf{v}})=\sqrt{(\bar{\mathbf{v}}^T\bar{\mathbf{v}})/d}$. 
\end{lemma}

\begin{lemma}[]\label{lem:derivative}
Consider any vector $\mathbf{v} \in \mathbb{R}^d$. Let $\bar{\mathbf{v}}$ and $\tilde{\mathbf{v}}$ be its mean-subtracted and standardized versions, respectively.
The derivative of $\tilde{\mathbf{v}}$ with respect to $\bar{\mathbf{v}}$ is then given by
\begin{align*}
\frac{\partial \tilde{\mathbf{v}}}{\partial \bar{\mathbf{v}}} &= \frac{\rho}{\sigma(\bar{\mathbf{v}})} \left( \mathbf{I} - \frac{1}{d(\sigma(\bar{\mathbf{v}}))^2} \bar{\mathbf{v}}\bar{\mathbf{v}}^T \right) \\
&= \frac{\rho}{\sigma(\bar{\mathbf{v}})} \left( \mathbf{I} - \frac{1}{\left\Vert \bar{\mathbf{v}} \right\Vert^2} \bar{\mathbf{v}}\bar{\mathbf{v}}^T \right) \\
&= \frac{\rho}{\sigma(\bar{\mathbf{v}})} \left( \mathbf{I} - \mathbf{P}_{\bar{\mathbf{v}}} \right). \numberthis \label{eq:lemma2_2}
\end{align*}
Also, based on Lemma~\ref{lem:mean}, the derivative of $\bar{\mathbf{v}}$ with respect to $\mathbf{v}$ is given by
\begin{align*}
\frac{\partial \bar{\mathbf{v}}}{\partial \mathbf{v}} &= \left( \mathbf{I} - \mathbf{P}_{\mathbf{1}} \right). \numberthis \label{eq:lemma2_3}
\end{align*}
Since $\sigma(\bar{\mathbf{v}}) = \sigma(\mathbf{v})$, by the chain rule, we can derive the gradient of a loss function $\mathcal{L}$ with respect to $\mathbf{v}$ as follows:
\begin{align}
\frac{\partial \mathcal{L}}{\partial \mathbf{v}} &= 
\frac{\rho}{\sigma(\bar{\mathbf{v}})} 
\left( \mathbf{I}-\mathbf{P}_{\mathbf{1}} \right) 
\left( \mathbf{I}-\mathbf{P}_{\bar{\mathbf{v}}} \right)
\frac{\partial \mathcal{L}}{\partial \tilde{\mathbf{v}}}.
\end{align}
\end{lemma}
\clearpage

\section{Derivation of Quantization Errors}
\label{supp:deriv2}
In this section, we derive the expected quantization error, which measures the difference between the original LMPUs and their quantized values, where $p(x)$ represents a standard normal distribution. The error is formulated as
\begin{align}
\mathbb{E} \left[ (\Delta w - \Delta \bar{w} \right)^2] &= 
\sum_{r=0}^{R} \int_{u_r}^{u_{r+1}} (x - q_r)^2 p(x) dx
\end{align}
where $q_r$ is the quantization level. To evaluate the integral, we expand the squared term as follows:
\begin{align}
\int (x - q_r)^2 p(x) dx &= \frac{1}{\sqrt{2\pi}} \int (x - q_r)^2 e^{-\frac{x^2}{2}} dx \nonumber \\
&= \frac{1}{\sqrt{2\pi}} \left( \underbrace{\int x^2 e^{-\frac{x^2}{2}} dx}_{P_1} \underbrace{- 2q_r \int x e^{-\frac{x^2}{2}} dx}_{P_2} + \underbrace{q_r^2 \int e^{-\frac{x^2}{2}} dx}_{P_3} \right)
\end{align}
We now calculate each term $P_1$, $P_2$, and $P_3$.
Let $t=\frac{x}{\sqrt{2}}$, which transforms $P_1$ into
\begin{align}
P_1 &= 2\sqrt{2} \int t^2 e^{-t^2} dt \nonumber \\
&= -\sqrt{2} t e^{-t^2} + \sqrt{2} \int e^{-t^2} dt \quad \left( \because \int udv = uv - \int v du \quad \text{where } u=t \text{ and } dv=te^{-t^2}dt \right)
\end{align}
The definite integral over the quantization boundaries is then given by
\begin{align}
-\sqrt{2} \left[ t e^{-t^2} \right]_{u_r / \sqrt{2}}^{u_{r+1} / \sqrt{2}} + \sqrt{2} \int_{u_r / \sqrt{2}}^{u_{r+1} / \sqrt{2}} e^{-t^2} dt  &= \left( u_r e^{-\frac{u_r^2}{2}} - u_{r+1} e^{-\frac{u_{r+1}^2}{2}} \right) + \sqrt{\frac{\pi}{2}} \left( \textrm{erf}\left( \frac{u_{r+1}}{\sqrt{2}} \right) - \textrm{erf}\left( \frac{u_{r}}{\sqrt{2}} \right) \right)
\end{align}
Also, we can evaluate the definite integral of $P_2$ over the qunatization boundaries as follows:
\begin{align}
-2q_r \int_{u_r}^{u_{r+1}} x e^{-\frac{x^2}{2}} dx &= 2q_r \left[ e^{-x^2} \right]_{u_r}^{u_{r+1}} \quad \quad \quad \left( \because \int x e^{-\frac{x^2}{2}} dx = -e^{-\frac{x^2}{2}} \right)\nonumber \\
&= 2q_r (e^{-\frac{u_{r+1}^2}{2}} - e^{-\frac{u_{r}^2}{2}})
\end{align}
Finally, we can easily obtain the definite integral of $P_3$ over the qunatization boundaries by substituting $t=\frac{x}{\sqrt{2}}$, as follows:
\begin{align}
q_r^2 \int_{u_r}^{u_{r+1}} e^{-\frac{x^2}{2}} dx &= \sqrt{2} q_r^2 \int_{u_r/\sqrt{2}}^{u_{r+1} / \sqrt{2}} e^{-t^2} dt \nonumber \\
&= \sqrt{\frac{\pi}{2}} q_r^2 \left( (\textrm{erf}\left( \frac{u_{r+1}}{\sqrt{2}} \right) - \textrm{erf}\left( \frac{u_{r}}{\sqrt{2}} \right)\right)
\end{align}
Combining all the above derivations and unrolling the sum, we can obtain the final expression for the expected quantizaion error as follows:
\begin{align}
\sum_{r=0}^{R} \int_{u_r}^{u_{r+1}} (x - q_r)^2 p(x) dx 
&= \sum_{r=0}^{R} \bigg\{ \frac{1}{\sqrt{2\pi}} (2q_r - u_{r+1}) e^{-\frac{u_{r+1}^2}{2}} - \frac{1}{\sqrt{2\pi}} (2q_r - u_r) e^{-\frac{u_{r}^2}{2}} \nonumber\\
& \quad \quad \quad+ \frac{1}{2} (q_r^2 + 1) \left( \mathrm{erf}\left( \frac{u_{r+1}}{\sqrt{2}} \right) -\mathrm{erf}\left( \frac{u_{r}}{\sqrt{2}} \right) \right)\bigg\} \nonumber \\
&= \frac{1}{2}(q_{R}^2+1) 
-\sqrt{\frac{2}{\pi}} q_1 e^{-\frac{u_1^2}{2}}
-\frac{1}{2} q_1^2 \textrm{erf}\left( \frac{u_1}{\sqrt{2}} \right) \nonumber \\
& \quad + \sqrt{\frac{2}{\pi}} \sum_{r=1}^{R-1} (q_r-q_{r+1}) e^{-\frac{u_{r+1}^2}{2}} \nonumber + \frac{1}{2} \sum_{r=1}^{R-1} (q_r^2-q_{r+1}^2)\textrm{erf}\left( \frac{u_{r+1}}{\sqrt{2}} \right). \numberthis
\label{eq:cost_func}
\end{align}

\section{Experimental setup}
\label{supp:exp_setup}

\paragraph{Implementation details}
We follow most of the implementation setups and evaluation protocols in~\cite{kim2024fedacg,acar2021federated,xu2021fedcm,seo2024relaxed}.
The ResNet-18 architecture~\cite{he2016deep} is adopted as our backbone network. Consistent with~\citep{hsieh2020non} and common practice in FL, all BN layers in ResNet-18 are replaced with GN layers. 
Following the recommendation of Qiao~\etal~\citep{qiao2019ws}, WS is applied before each GN layer.
%This also follows the recommendation of Qiao~\etal~\cite{qiao2019ws} to use WS in combination with BN or GN layers for optimal performance. 
All the models are trained from scratch by using the SGD optimizer with an initial learning rate of 0.1 and a weight decay of 0.001. For the proposed model, the learning rate is exponentially decayed at each communication round by a factor of 0.995. For the other models compared, we select the learning decay parameter from $\{0.995, 0.998, 1\}$ to attain the best accuracy. The global learning rate of FedAdam is set to 0.01, and that of the other methods is set to 1. Momentum is not used following the previous works~\cite{kim2024fedacg,acar2021federated,xu2021fedcm}, and gradient clipping is applied for learning stability. Unless otherwise noted, the number of local training epochs per round is set to 5, with the batch size adjusted so that each local epoch consists of 10 iterations. The hyper-parameter of WS is set to $\rho=0.001$. The source code is implemented by using the PyTorch framework~\cite{paszke2019pytorch} on NVIDIA RTX 4090 GPUs. 
We set the number of local training epochs to 5. The batch size for local updates is adjusted so that each local epoch has 10 iterations (\ie, 50 iterations during a single communication round).

\paragraph{Hyper-parameter selection}

We adopt the hyper-parameter settings of the baseline methods suggested in~\cite{kim2024fedacg,seo2024relaxed}.
Table~\ref{table:hyper} summarizes the hyper-parameter settings we used, with the notations consistent with the original papers.

\begin{table}[h]
  \caption{Summary of hyper-parameter selection}
  \centering
  \resizebox{0.4\linewidth}{!}{
  \begin{tabular}{@{}c|l@{}} 
  \toprule
   Method & Hyper-parameters \\
   \midrule
   FedProx~\cite{li2020federated} & $\mu = 0.001$ \\ 
   FedAvgM~\cite{hsu2019measuring} & $\beta = 0.4$ \\
   FedADAM~\cite{reddi2021adaptive} & $\tau = 0.001$, $\beta_1 = 0.9$, $\beta_2 = 0.99$ \\
   FedDyn~\cite{acar2021federated} & $\alpha = 0.1$ \\
   FedMLB~\cite{kim2022multi} & $\tau = 1$, $\lambda_1 = 1$, $\lambda_2 = 1$ \\
   FedLC~\cite{zhang2022Federated} & $\tau = 1$ \\
   FedNTD~\cite{lee2022preservation} & $\tau = 1$, $\beta = 0.3$ \\
   FedDecorr~\cite{shi2022towards} & $\beta = 0.01$ \\
   FedRCL~\cite{seo2024relaxed} & $\tau = 0.05$, $\beta = 1$, $\lambda = 0.7$ \\
   FedACG~\cite{kim2024fedacg} & $\beta = 0.001$, $\lambda = 0.85$ \\
  \bottomrule
  \end{tabular}
}
\label{table:hyper}
\end{table}

\clearpage

\begin{figure}[h!]
\centering
\begin{subfigure}[b]{0.9\linewidth}
\centering
\includegraphics[width=1\linewidth]{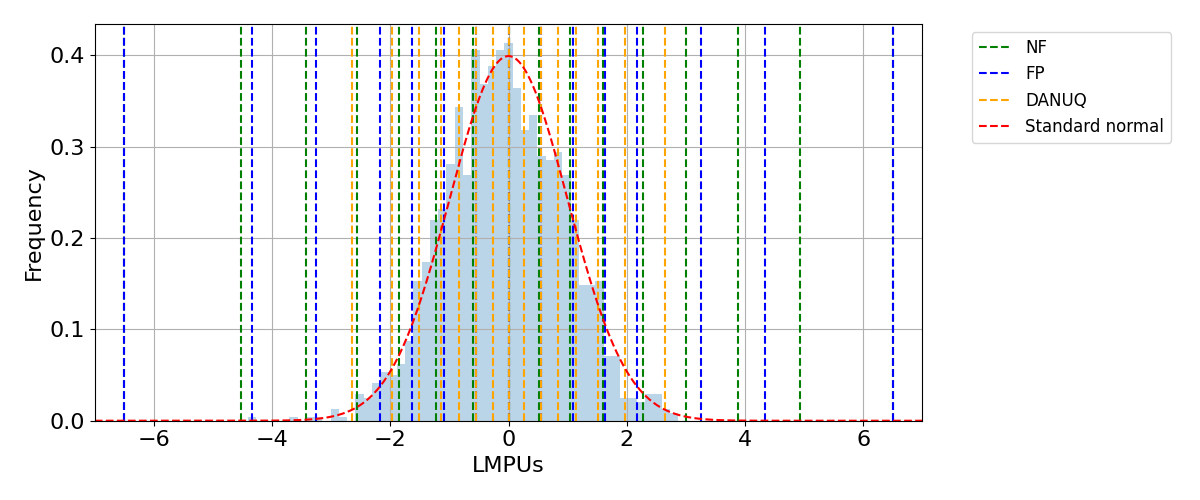}
\end{subfigure}
\begin{subfigure}[b]{0.9\linewidth}
\centering
\includegraphics[width=1\linewidth]{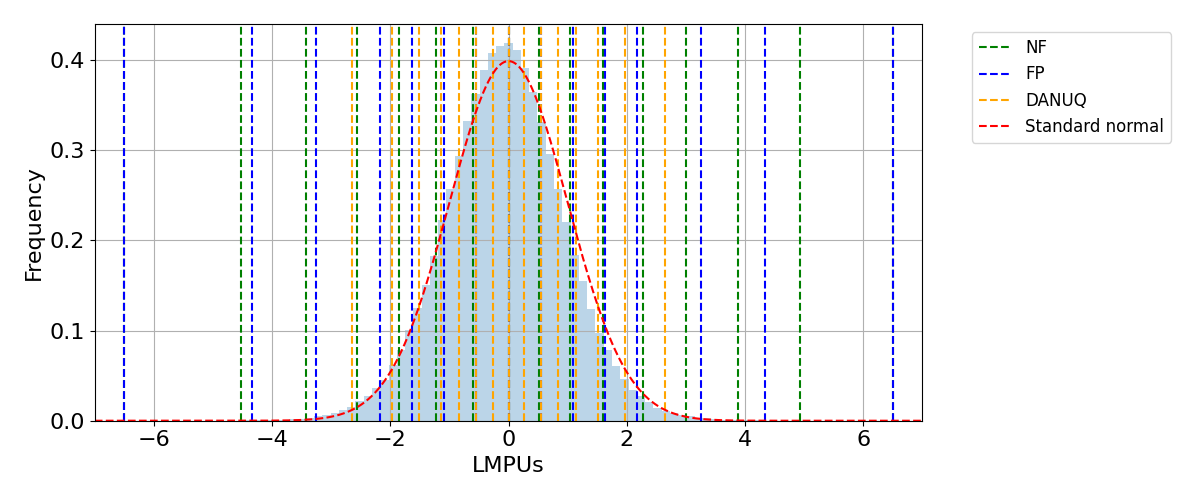}
\end{subfigure}
\hspace{0.4cm}
\caption{
Visualization of QLs used in different NUQ methods. The histogram represents the empirical distribution of LMPUs in the 1st and 3rd ResNet-18 blocks, and the red curve denotes the standard normal distribution. The vertical dashed lines indicate the QLs chosen by different methods, NF, FP, and the proposed DANUQ, where the 4-bit representation is used.
}
%\vspace{-0.8cm}
\label{fig:qls_explain}
\end{figure}

\paragraph{QLs of NUQ methods} Figure~\ref{fig:qls_explain} provides a comparative visualization of the QLs adopted by different NUQ methods. The histogram illustrates the empirical distribution of LMPUs with the standard normal distribution curve.
As shown in the figure, the proposed DANUQ places QLs more adaptively based on the statistical structure of LMPUs, leading to improved quantization efficiency.

\clearpage

\section{Convergence plot evaluated on various federated learning scenarios}
\label{ssup:convergence_plot}
\begin{figure}[h!]
\centering
\begin{subfigure}[b]{0.48\linewidth}
\centering
\includegraphics[width=1\linewidth]{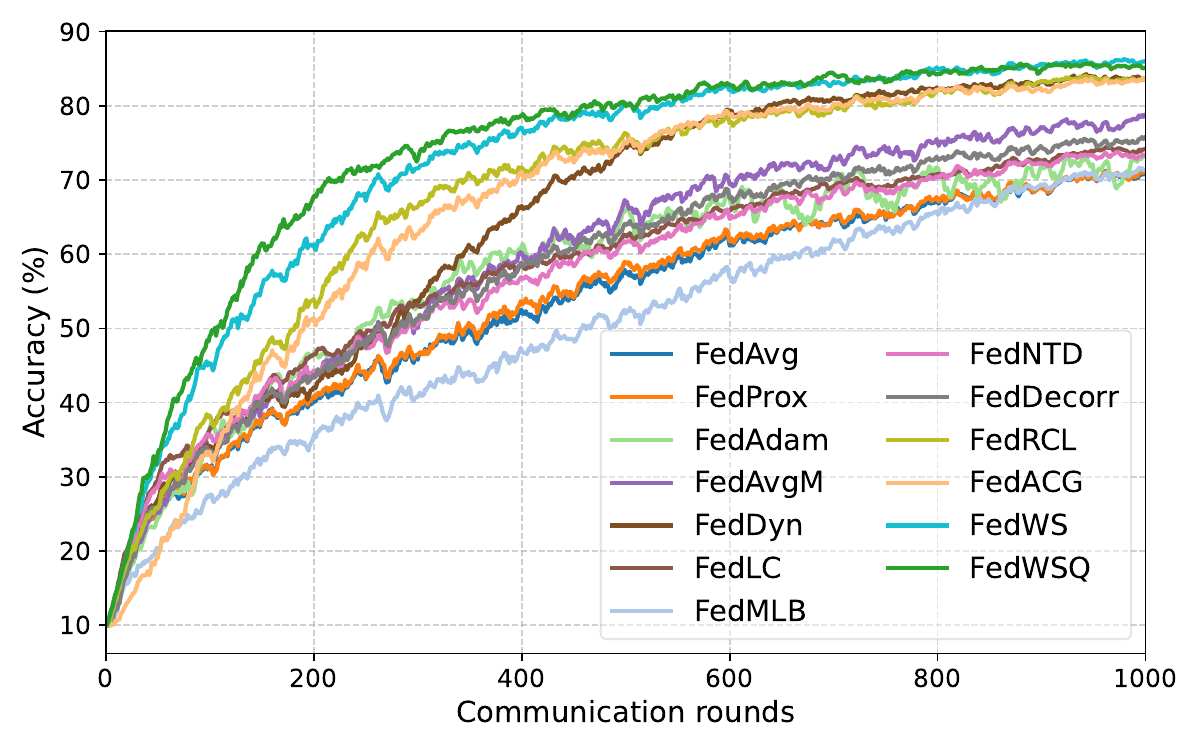}
%\vspace{-0.4cm}
\caption{$\alpha=0.1$, 5\% participation over 100 clients}
%\caption{1\% participation, 100 clients}
\end{subfigure}
\begin{subfigure}[b]{0.48\linewidth}
\centering
\includegraphics[width=1\linewidth]{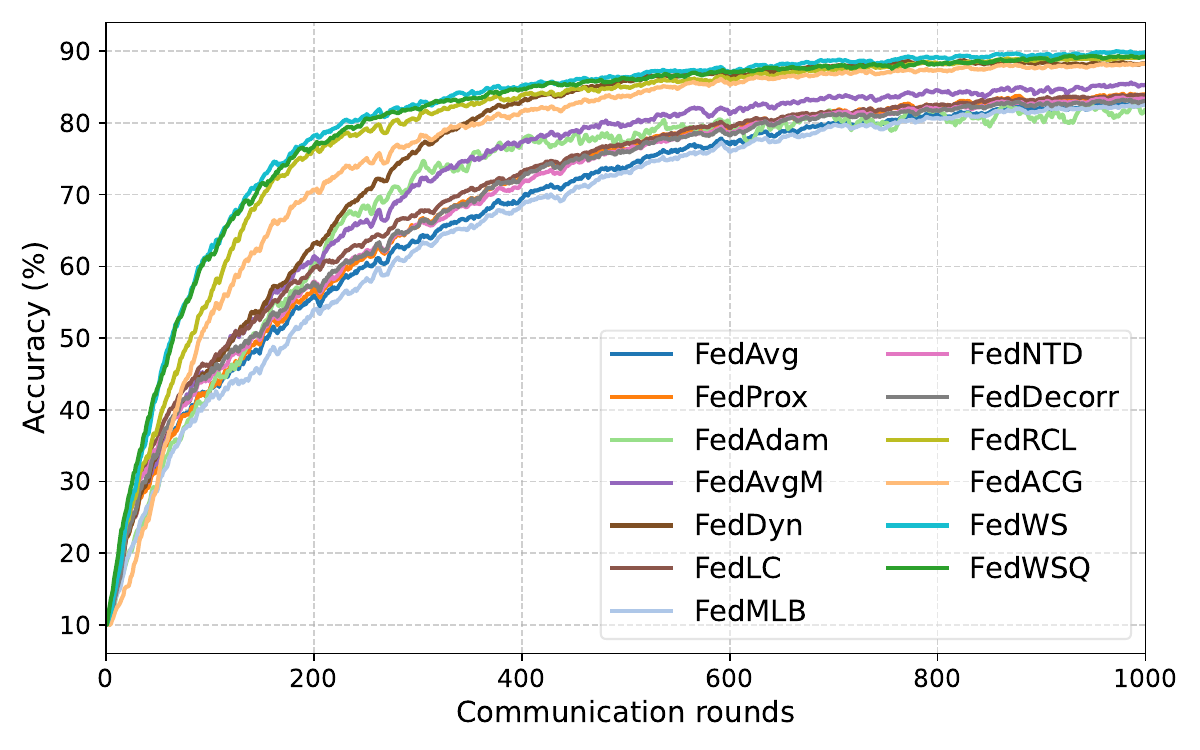}
%\vspace{-0.4cm}
\caption{$\alpha=0.3$, 5\% participation over 100 clients}
\end{subfigure}
%\rulesep
\hspace{0.4cm}

\begin{subfigure}[b]{0.48\linewidth}
\centering
\includegraphics[width=1\linewidth]{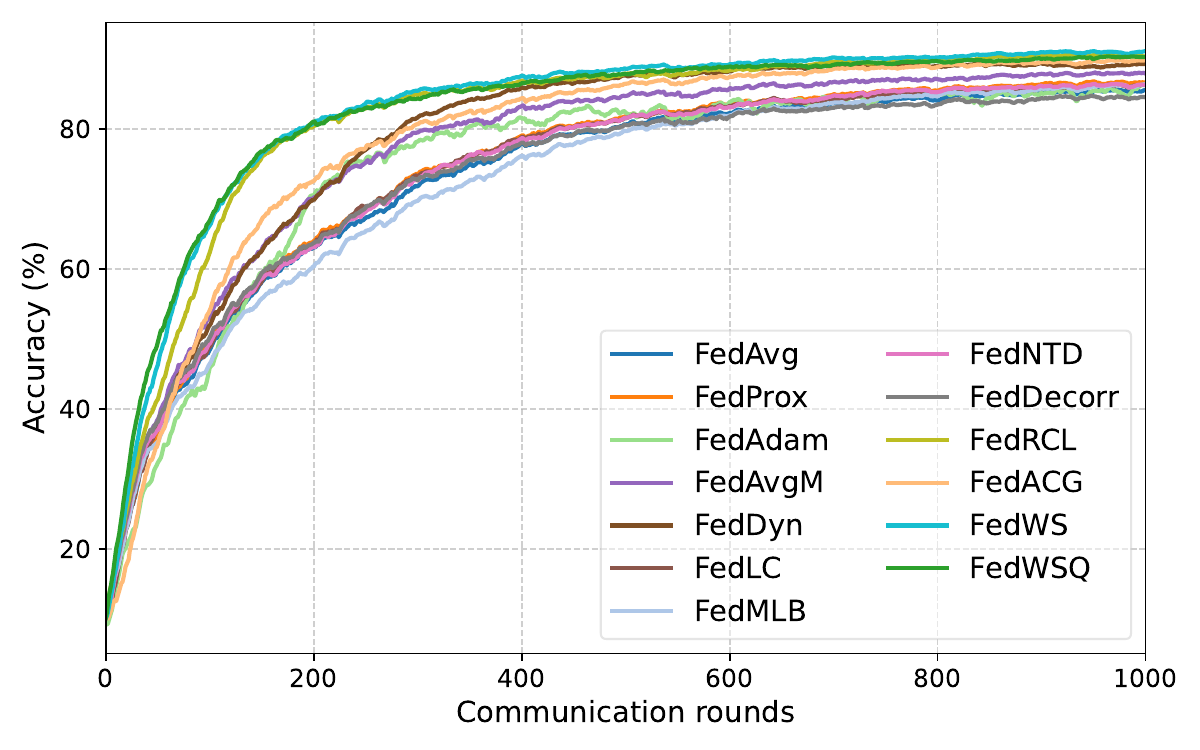}
%\vspace{-0.4cm}
\caption{$\alpha=0.6$, 5\% participation over 100 clients}
%\caption{1\% participation, 100 clients}
\end{subfigure}
\begin{subfigure}[b]{0.48\linewidth}
\centering
\includegraphics[width=1\linewidth]{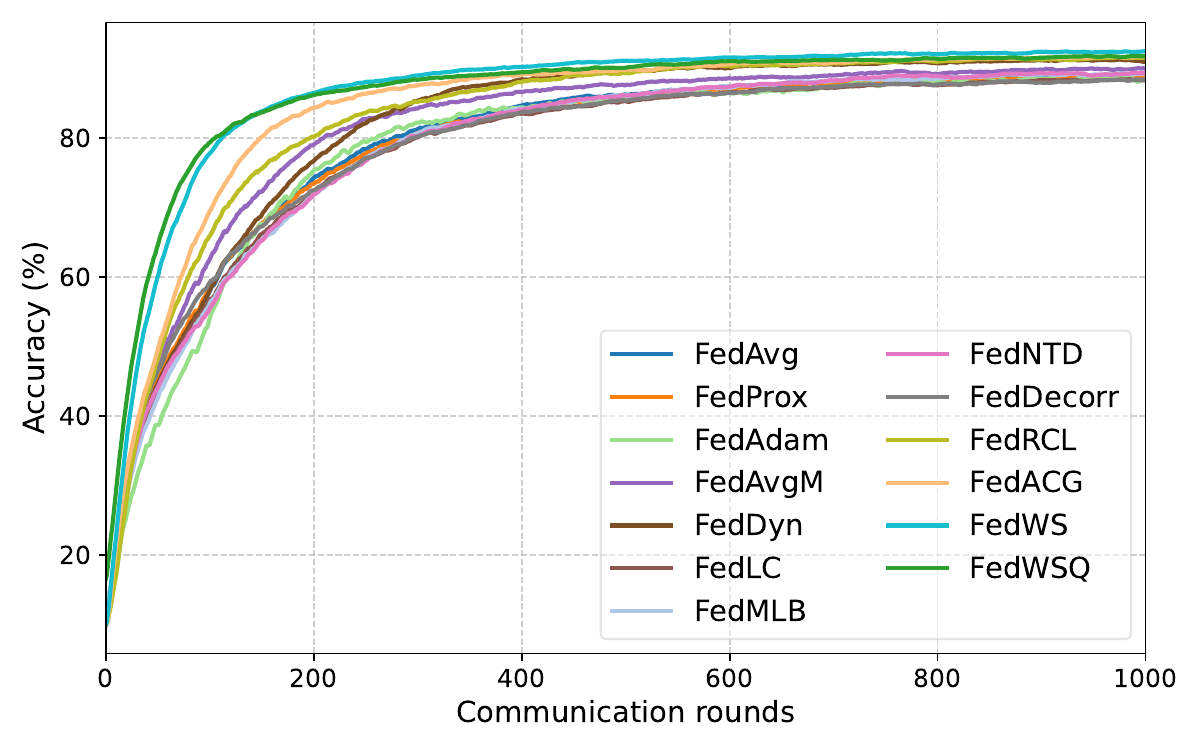}
%\vspace{-0.4cm}
\caption{\textit{i.i.d.}, 5\% participation over 100 clients}
\end{subfigure}
%\rulesep
\hspace{0.4cm}

\caption{
Convergence plots of our FedWS and FedWSQ compared to conventional methods on CIFAR-10 with 5\% participation over 100 clients under varying Dirichlet parameters.
}
%\vspace{-0.8cm}
\label{fig:convergence_cifar10}
\end{figure}

\begin{figure}[h!]
\centering
\begin{subfigure}[b]{0.48\linewidth}
\centering
\includegraphics[width=1\linewidth]{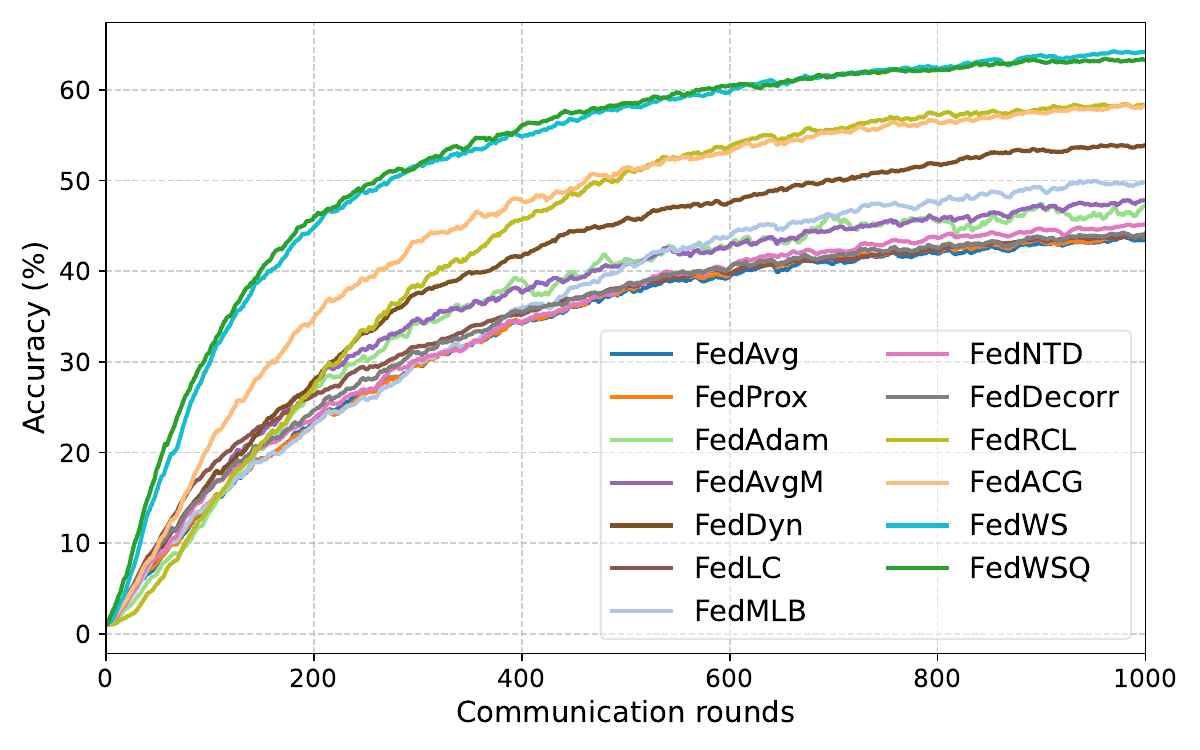}
%\vspace{-0.4cm}
\caption{$\alpha=0.1$, 5\% participation over 100 clients}
%\caption{1\% participation, 100 clients}
\end{subfigure}
\vspace{0.3cm}
\begin{subfigure}[b]{0.48\linewidth}
\centering
\includegraphics[width=1\linewidth]{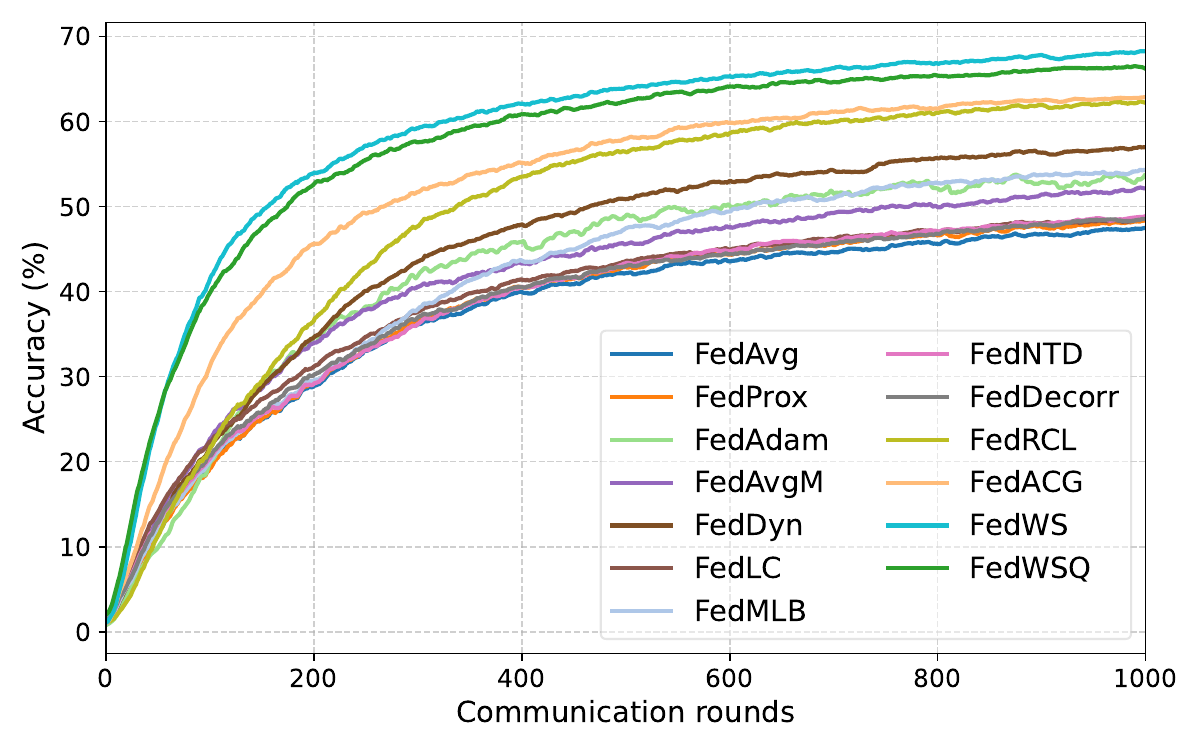}
%\vspace{-0.4cm}
\caption{$\alpha=0.3$, 5\% participation over 100 clients}
\end{subfigure}
%\rulesep
\hspace{0.4cm}

\begin{subfigure}[b]{0.48\linewidth}
\centering
\includegraphics[width=1\linewidth]{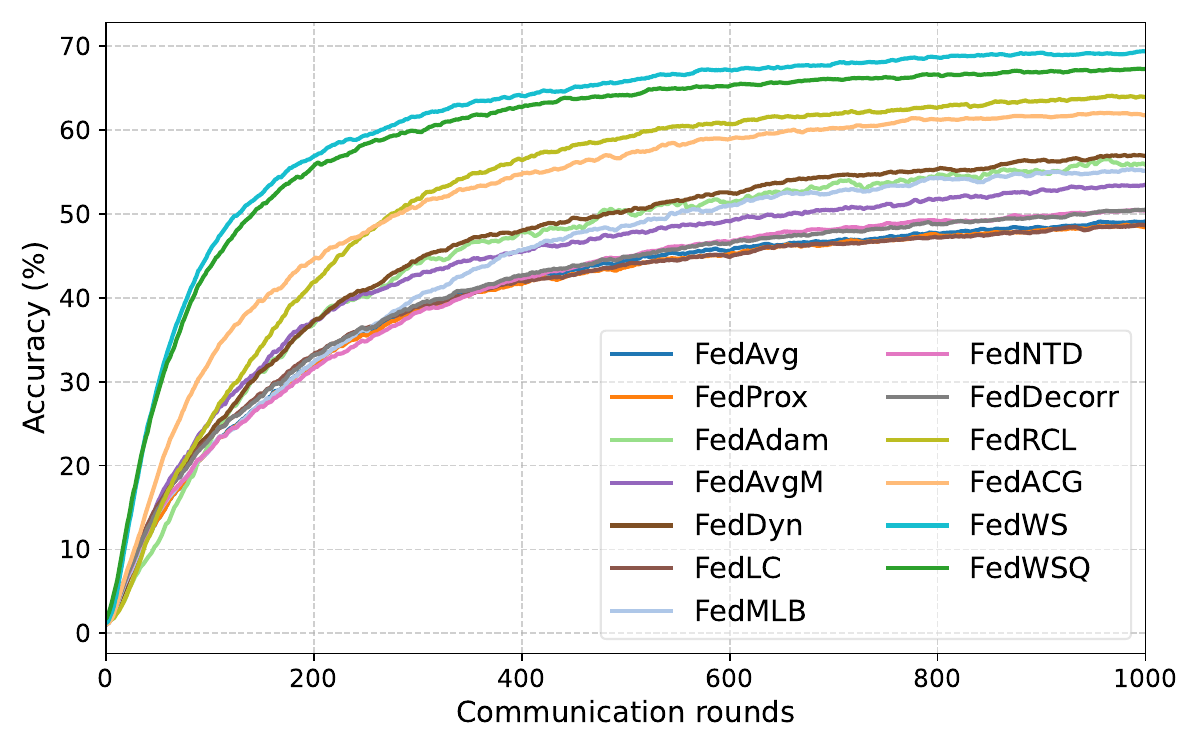}
%\vspace{-0.4cm}
\caption{$\alpha=0.6$, 5\% participation over 100 clients}
%\caption{1\% participation, 100 clients}
\end{subfigure}
\vspace{0.3cm}
\begin{subfigure}[b]{0.48\linewidth}
\centering
\includegraphics[width=1\linewidth]{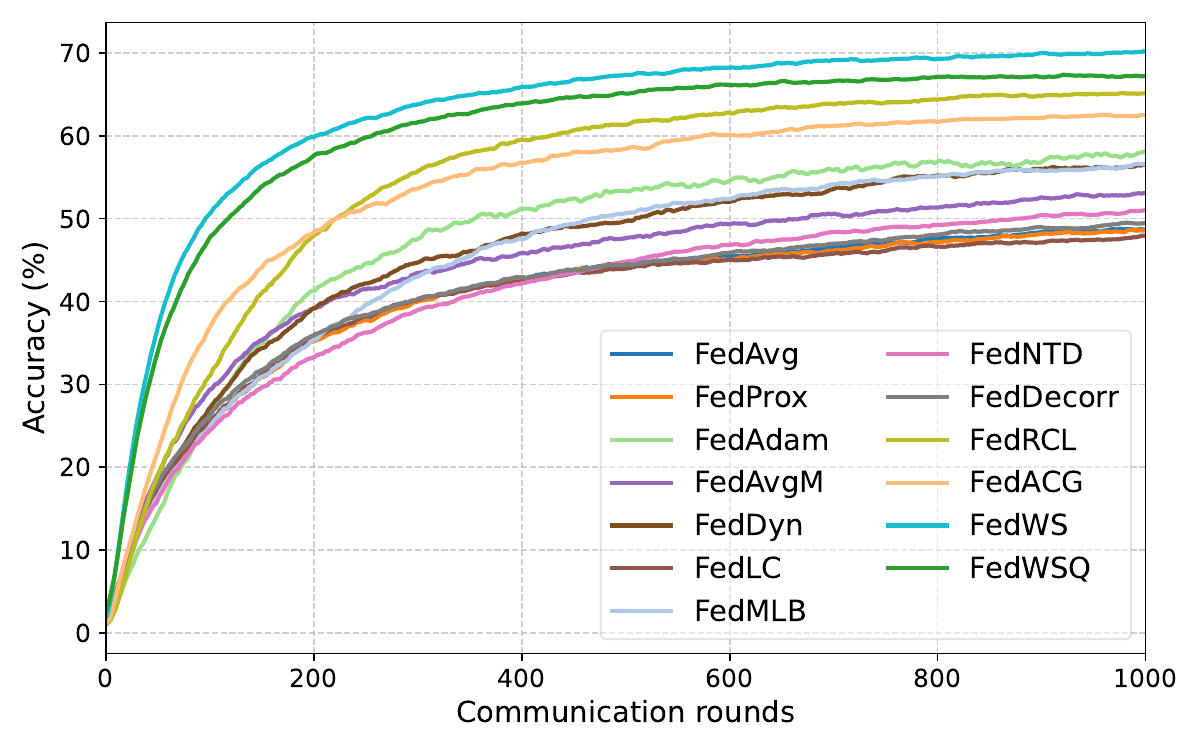}
%\vspace{-0.4cm}
\caption{\textit{i.i.d.}, 5\% participation over 100 clients}
\end{subfigure}
%\rulesep
\hspace{0.4cm}

\caption{
Convergence plots of our FedWS and FedWSQ compared to conventional methods on CIFAR-100 with 5\% participation over 100 clients under varying Dirichlet parameters.
}
%\vspace{-0.8cm}
\label{fig:convergence_cifar100}
\end{figure}

\begin{figure}[h!]
\centering
\begin{subfigure}[b]{0.48\linewidth}
\centering
\includegraphics[width=1\linewidth]{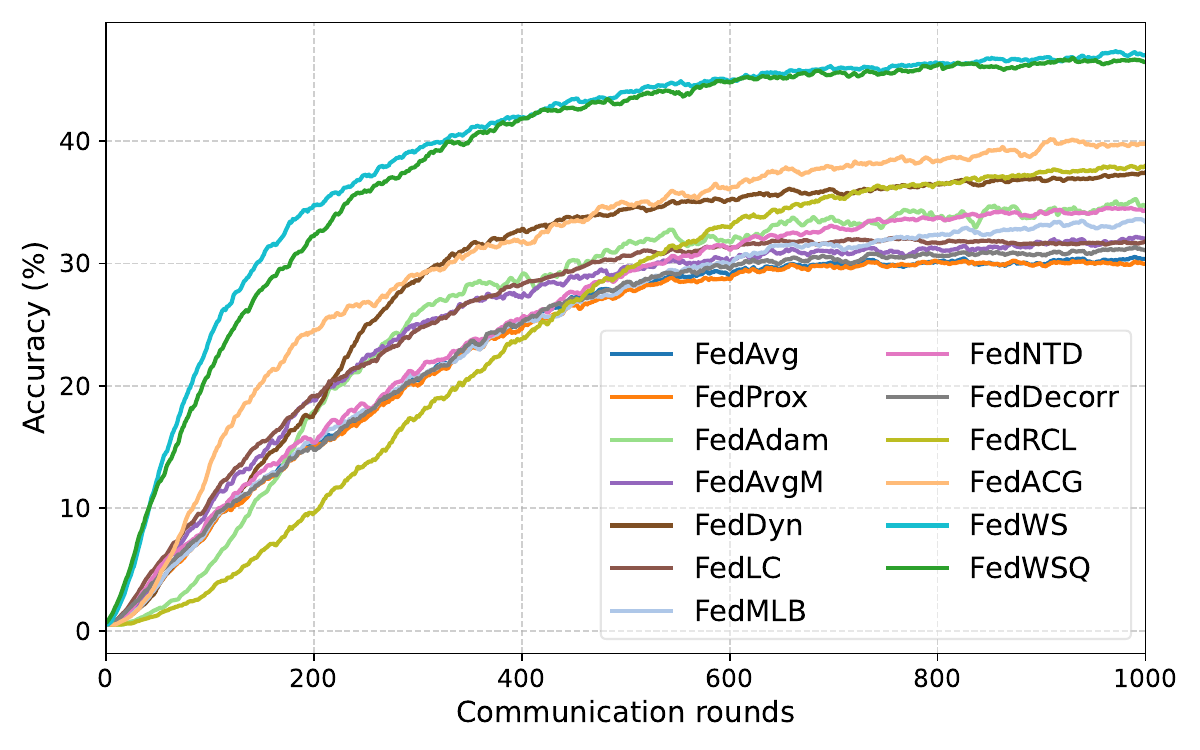}
%\vspace{-0.4cm}
\caption{$\alpha=0.1$, 5\% participation over 100 clients}
%\caption{1\% participation, 100 clients}
\end{subfigure}
\vspace{0.3cm}
\begin{subfigure}[b]{0.48\linewidth}
\centering
\includegraphics[width=1\linewidth]{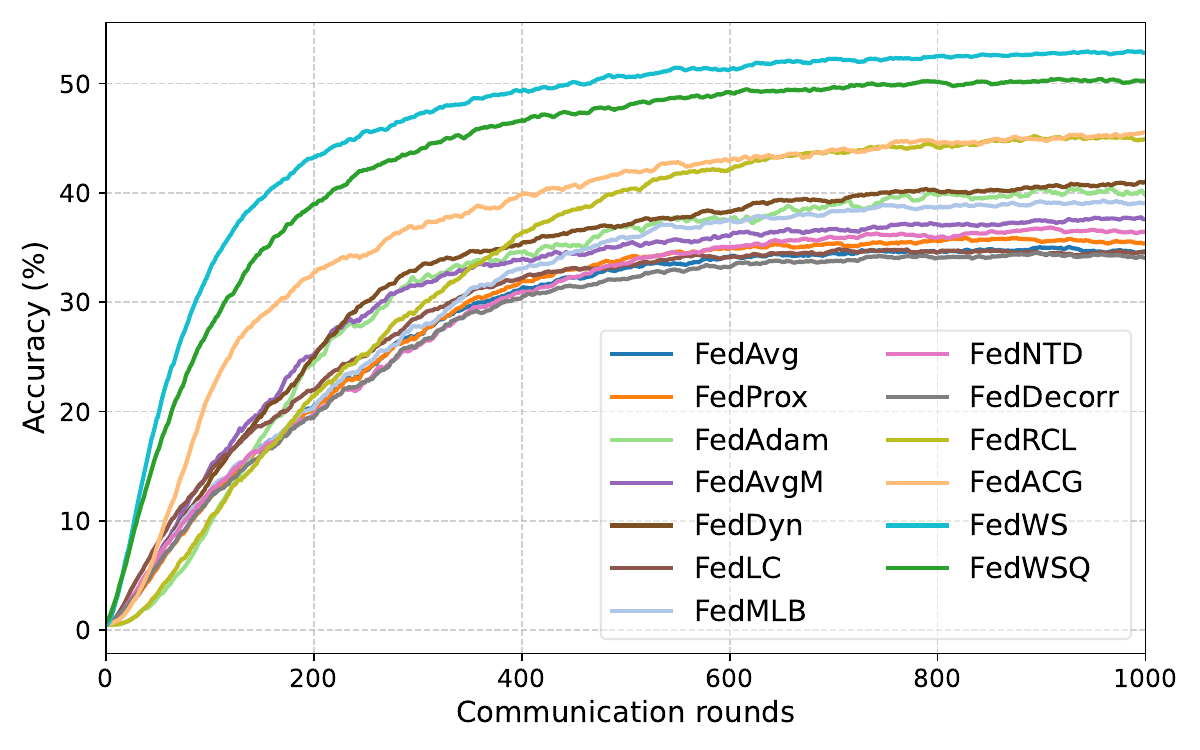}
%\vspace{-0.4cm}
\caption{$\alpha=0.3$, 5\% participation over 100 clients}
\end{subfigure}
%\rulesep
\hspace{0.4cm}

\begin{subfigure}[b]{0.48\linewidth}
\centering
\includegraphics[width=1\linewidth]{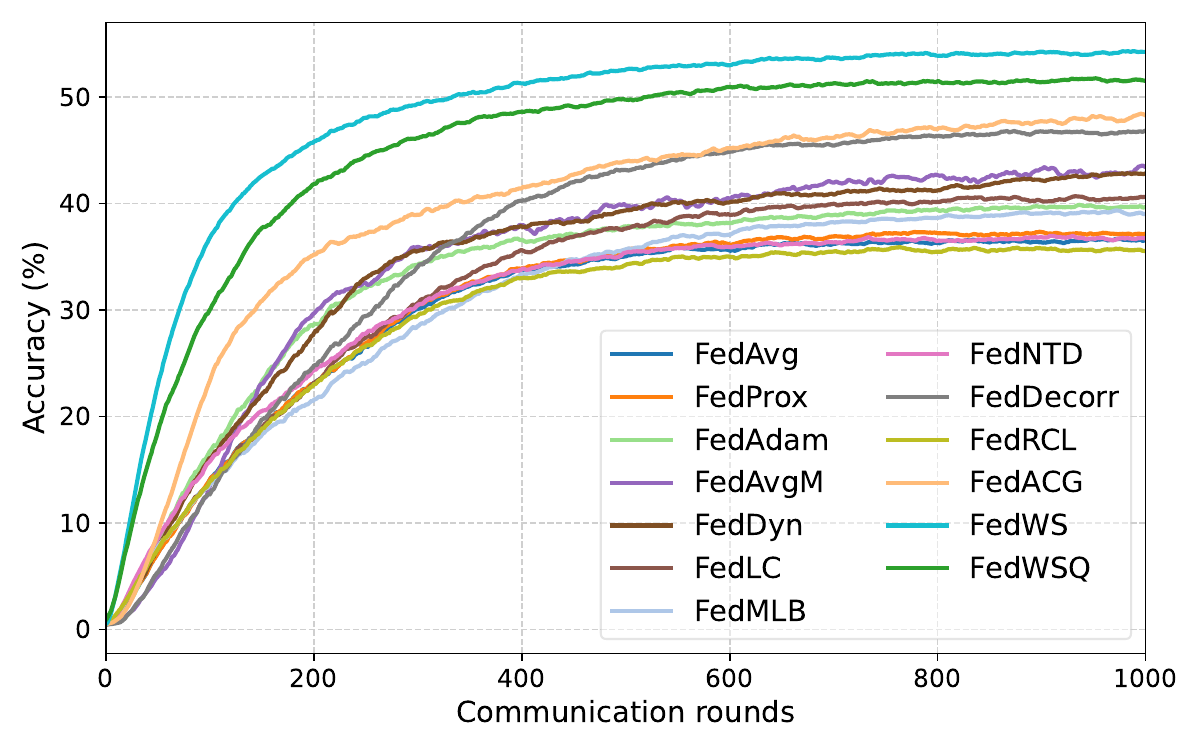}
%\vspace{-0.4cm}
\caption{$\alpha=0.6$, 5\% participation over 100 clients}
%\caption{1\% participation, 100 clients}
\end{subfigure}
\vspace{0.3cm}
\begin{subfigure}[b]{0.48\linewidth}
\centering
\includegraphics[width=1\linewidth]{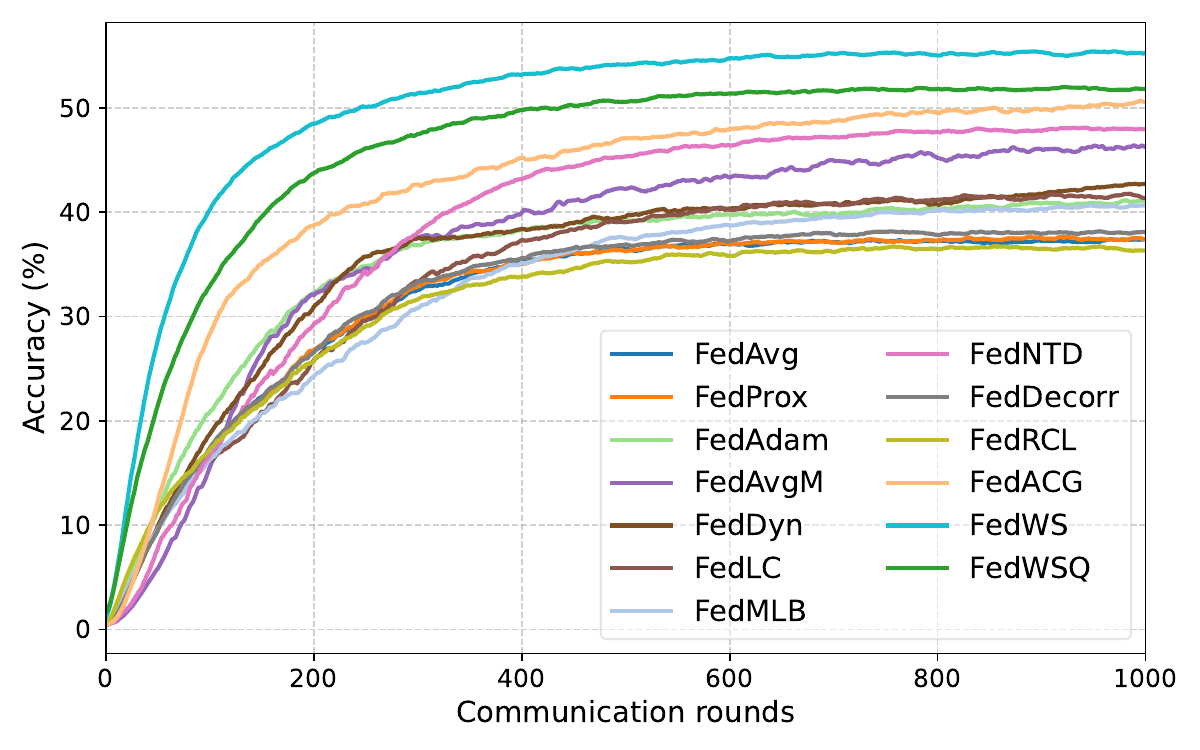}
%\vspace{-0.4cm}
\caption{\textit{i.i.d.}, 5\% participation over 100 clients}
\end{subfigure}
%\rulesep
\hspace{0.4cm}

\caption{
Convergence plots of our FedWS and FedWSQ compared to conventional methods on Tiny-ImageNet with 5\% participation over 100 clients under varying Dirichlet parameters.
}
%\vspace{-0.8cm}
\label{fig:convergence_tiny}
\end{figure}

Figures~\ref{fig:convergence_cifar10}-\ref{fig:convergence_tiny} present the convergence plots of various FL methods on CIFAR-10, CIFAR-100, and Tiny-ImageNet, for \textit{i.i.d} and non-\textit{i.i.d.} data distributions with $\alpha \in \{0.05, 0.1, 0.3, 0.6\}$, using a participation rate of 5\% over 100 distributed clients. 
As shown in the figures, FedWSQ consistently enhances the FL performance of conventional methods, outperforming those of state-of-the-art FL approaches.

%%%%%%%%%%%%%% CIFAR-10 %%%%%%%%%%%%%%%%%%%%%%%%%%%%%%%%%%%%%%%%%%%%%%%%%%%%%
%%%%%%%%%%%%%%%%%%%%%%%%%%%%%%%%%%%%%%%%%%%%%%%%%%%%%%%%%%%%%%%%%%%%%%%%%%%%%%%%%%%%%%%%%%%%%%%%%%%%%%%%%%%%%%%

\clearpage

%%%%%%%%%%%%%% CIFAR-100 %%%%%%%%%%%%%%%%%%%%%%%%%%%%%%%%%%%%%%%%%%%%%%%%%%%%%
%%%%%%%%%%%%%%%%%%%%%%%%%%%%%%%%%%%%%%%%%%%%%%%%%%%%%%%%%%%%%%%%%%%%%%%%%%%%%%%%%%%%%%%%%%%%%%%%%%%%%%%%%%%%%%%

\clearpage

%%%%%%%%%%%%%% Tiny-ImageNet %%%%%%%%%%%%%%%%%%%%%%%%%%%%%%%%%%%%%%%%%%%%%%%%%%%%%
%%%%%%%%%%%%%%%%%%%%%%%%%%%%%%%%%%%%%%%%%%%%%%%%%%%%%%%%%%%%%%%%%%%%%%%%%%%%%%%%%%%%%%%%%%%%%%%%%%%%%%%%%%%%%%%

\clearpage

\end{document}